\documentclass[final,1p,times,authoryear]{elsarticle}
\makeatletter
\def\ps@pprintTitle{%
    \let\@oddhead\@empty
    \let\@evenhead\@empty
    \def\@oddfoot{\footnotesize\itshape
        {Preprint submitted to an international journal} \hfill\today}%
    \let\@evenfoot\@oddfoot
    }
\makeatother
\usepackage{multicol}
\usepackage{multirow}
\usepackage{tabularx}
\usepackage{booktabs}

\usepackage{subcaption}
\usepackage[draft]{changes}   
\usepackage[normalem]{ulem}   
\setaddedmarkup{{#1}}
\setdeletedmarkup{{\sout{#1}}}
\usepackage{threeparttable} 
\colorlet{red}{black}

\usepackage{amssymb}
\usepackage{amsmath}
\usepackage{array}
\usepackage{float}
\usepackage{makecell}
\journal{Transportation research part C}

\begin{document}

\begin{frontmatter}



\title{Incorporating Graph Neural Networks into Route Choice Models}

\author[label1,label2]{Ma Yuxun}
\affiliation[label1]{organization={Department of Civil and Environmental Engineering, Institute of Science Tokyo
},
            addressline={2-12-1-M6-10},
            city={Meguro,Ookayama},
            postcode={152-8552},
            state={Tokyo},
            country={Japan}}

\author[label1]{Toru Seo} 


\begin{abstract}
{
Route choice models are one of the most important foundations for transportation research. 
Traditionally, theory-based models have been utilized for their great interpretability, such as logit models and Recursive Logit models. Recently, hybrid models that incorporate machine learning into discrete choice models have gained increasing attention. However, applying such hybrid models to route choice problem in general and large-scale road networks remains challenging. 
{In this study, hybrid models integrating the Recursive Logit model with Graph Neural Networks (GNNs) are proposed to improve prediction performance while partially preserving behavioral interpretability. The proposed models constitutes the interpretable systematic component, while the GNN serves as a flexible residual component. As in other hybrid choice models, the residual component improves prediction accuracy at the cost of behavioral transparency.}
To the authors' knowledge, GNNs have not been utilized for route choice modeling, despite their proven effectiveness in capturing road network features and their widespread use in other transportation research areas. The use of GNN can also relax the Independence of Irrelevant Alternatives property without relying on strong assumptions. The proposed model can also better capture the substitution pattern between two paths. By applying the proposed models to large-scale travel trajectory data from Tokyo with time-varying network features, the results demonstrate that they achieve higher prediction accuracy compared with existing models under both a standard chronological data split and additional temporal and destination holdout experiments.}
\end{abstract}




\end{frontmatter}



\section{Introduction}
\label{sec1}
\def\({\left(}
\def\){\right)}

Understanding and predicting how travelers choose routes based on road network characteristics is essential for transportation research and practice. With advancements in positioning systems, there is now an increasing amount of travel trajectory data available for route choice modeling. The increased quantity of such data has also enabled the application of big data techniques in this research area.

The most common approach for route choice modeling is the logit model \citep{ben2004route}, and its extensions such as the Recursive Logit model (RL) \citep{fosgerau2013link}. They are based on the Random Utility Maximization (RUM) theory, which assumes that travelers consistently choose routes that maximize their expected utility based on given utility functions. These theory-based route choice models offer high interpretability, as they assume that the utility function has a predefined and interpretable form and their coefficients can be assigned clear behavioral or economic meanings. For example, the value of time can be inferred from the weight of the time component in a linear utility function that includes time and other route attributes. However, standard logit models face several challenges. First, their utility functions are predefined, meaning that the models' accuracy heavily depends on the modeler's prior knowledge. This constraint can hinder the model's ability to capture the complex and heterogeneous nature of real-world route choice behavior. Second, these models exhibit the Independence of Irrelevant Alternatives (IIA) property and cannot capture the substitution effects \footnote{Substitution effects in choice problem means how the choice probability of a changed or removed alternative is redistributed among the remaining alternatives.}.

Recent studies have begun to integrate logit-type models with deep learning models to capture complex relationships among variables while maintaining the output rationality of discrete choice models \citep{van2022choice,han2020neural,sifringer2020enhancing,wong2021reslogit,phan2022attentionchoice}. These hybrid models have shown a promising balance between prediction accuracy and interpretability\footnote{Interpretability may have different meanings across models. For example, \citet{wong2021reslogit} show that ResLogit retains interpretability comparable to that of the Multinomial Logit model while using its residual component to capture cross-effects. Similarly, in this study, interpretability is provided by the systematic component, whose coefficients retain clear economic meanings by linking link features to choice outcomes.}, particularly in mode choice analysis and other transportation applications. Among these hybrid models, the most common approach is to divide utility into a knowledge-driven component and a data-driven component. The knowledge-driven component is specified based on prior knowledge, such as how factors affect human behavior. The data-driven component is used to capture unobserved heterogeneity across alternatives.

However, it is challenging to directly apply these hybrid models to route choice modeling in large-scale road networks for three main reasons. First, these hybrid models typically rely on a pre-generated candidate path set, whereas in route choice modeling the choice set construction can substantially affect estimation results \citep{bekhor2006evaluation}. In theory-based model, this issue has been resolved by the RL; however, extending RL to hybrid model is not trivial. Second, it is difficult to extract and encode network-based spatial characteristics as effective inputs for data-driven components. Third, selecting a suitable model to capture heterogeneous dependencies between paths under different spatial characteristics remains challenging.

One of the most significant advancements in ML domain is Graph Neural Networks (GNNs), which can capture spatial dependencies between nodes. Since a road network can be naturally represented by a graph, GNNs are applied to problems in transportation research, including traffic flow prediction \citep{he2023stgc,bai2020adaptive,song2020spatial}, travel time prediction \citep{lu2019leveraging,xie2020deep} and signal control \citep{zhong2021probabilistic, hu2020traffic}. {However, to the best of the authors’ knowledge, no existing study has incorporated GNNs into route choice modeling. Most ML-based route choice models adopt path-based formulations, which do not directly exploit the link-level topology of road networks, while existing link-based approaches mainly rely on DNNs or deep inverse reinforcement learning. \citep{zhao2023deep,ZHANG2024103485,qiu2025generalized}.}

{The objective of this paper is to extend the Recursive Logit (RL) framework by developing a hybrid link-based route choice model for large-scale networks that improves prediction performance while partially preserving the economic interpretability of RL. The utility function consists of a systematic component and a residual component. The systematic component provides behavioral interpretability through predefined utility functions, whereas the residual component improves prediction accuracy by capturing effects that are not represented by the systematic component.} To this end, the proposed model relaxes the IIA property and integrates Graph Neural Networks (GNNs) to better capture substitution effects between paths without imposing a predefined substitution structure. Similar to ResLogit \citep{wong2021reslogit}, the formulation is inspired by the Mother Logit model \citep{mcfadden1977application} and does not rely on a conventional random utility maximization (RUM) structure. The effectiveness of the proposed model is evaluated using a real-world travel trajectory dataset.

The contributions of this paper can be summarized as follows:

\begin{itemize}

\item A hybrid link-based route choice model, named {\it ResDGCN-RL}, is proposed by integrating RL with GNNs to capture road-network characteristics and learn cross-effects between actions. To the best of the authors' knowledge, this is the first attempt to incorporate GNNs into route choice modeling.

\item The proposed model not only relaxes the IIA property but also captures cross-effects between paths more effectively.

\item  An exogenous parameter is introduced into the loss function to explicitly control the trade-off between interpretability and prediction accuracy. 
\end{itemize}

The remainder of this paper is organized as follows. Section 2 reviews both logit-type and data-driven approaches to route choice modeling. Section 3 describes the formulation and key properties of the proposed model. {Section 4 demonstrates the mechanisms of proposed models by a toy network.} Section 5 evaluates their performance on real-world vehicle trajectory data collected in Tokyo. Finally, Section 6 concludes the paper and outlines directions for future research.

\section{Literature review}
\label{sec2}
\begin{figure}[t]
  \centering
  \includegraphics[width=\textwidth]{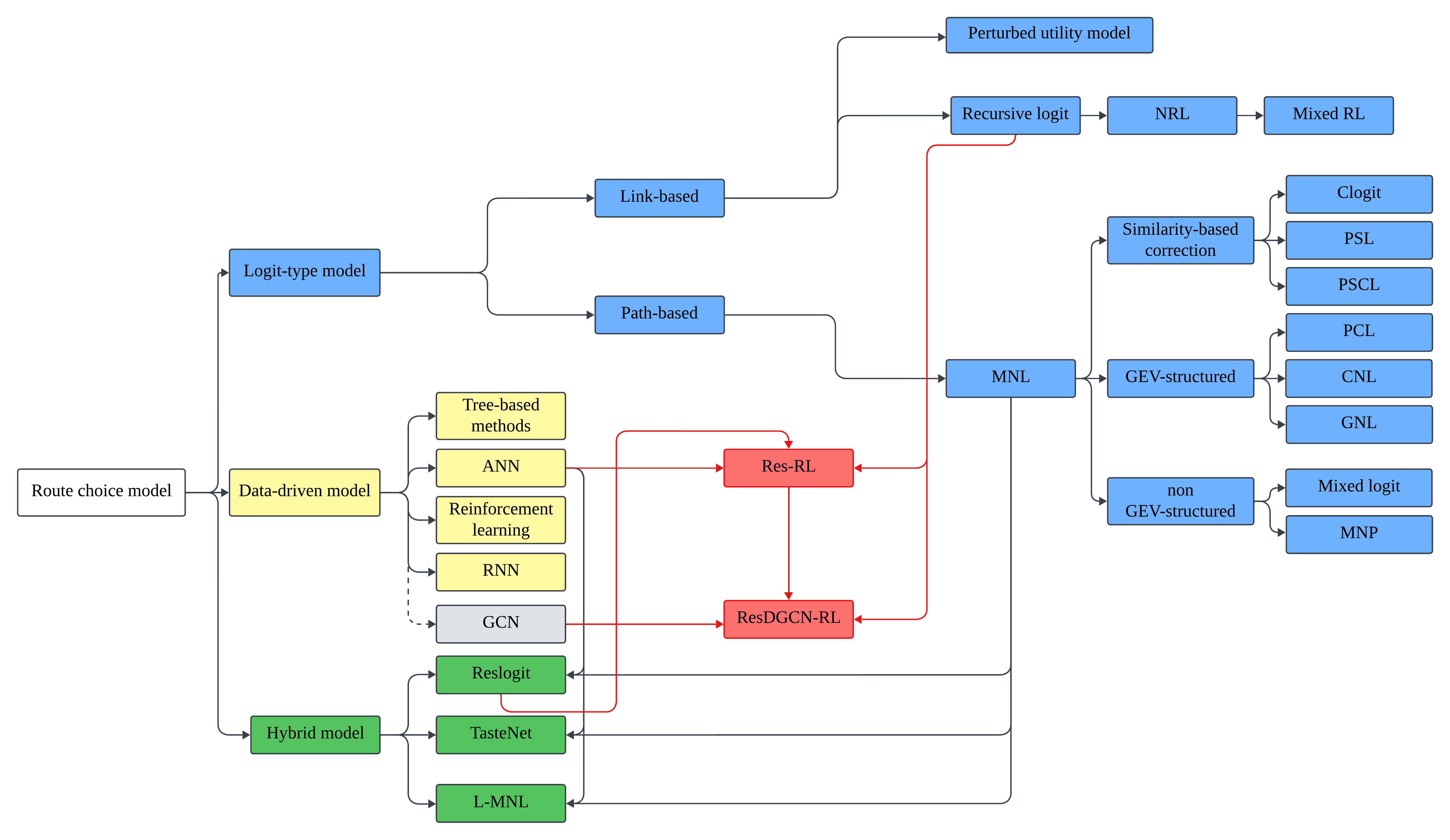}
  \caption{Conceptual framework of the literature review: research gaps and study positioning. This framework highlights existing research gaps, indicated by dotted lines, and emphasizes the focus of this study, represented by red lines and blocks.}
  \label{frameworkc2}
\end{figure}

Building on the discussion in the introduction, this section reviews existing literature on logit-based route choice models, data-driven route choice models, and hybrid route choice models. Figure \ref{frameworkc2} outlines the conceptual framework of the literature review, highlighting the major research gaps and clarifying the positioning of the present study.

\subsection{Logit-type models}

Discrete choice models are widely used in transportation research to analyze and predict individual route choice behavior among a finite set of path alternatives. Among these, logit-type route choice models have become particularly popular due to their analytical tractability and empirical reliability. The logit model is grounded in the framework of RUM, where the utility of each route is represented as the sum of a systematic component—based on observable characteristics of both the traveler and the route—and a random error term that accounts for unobservable factors, such as individual heterogeneity and latent preferences. Under this approach, it is assumed that travelers select the route with the highest total utility.

In logit-type route choice models, the random error terms are typically assumed to be independent and identically distributed (i.i.d.), which gives rise to the Independence of Irrelevant Alternatives (IIA) property. However, in route choice problems, different routes often share overlapping segments, introducing correlation among alternatives that violates the i.i.d. assumption. As a result, the IIA property can lead to unrealistic substitution patterns and biased estimates in the presence of overlapping routes.

To address the IIA limitation of the logit model, several similarity-based extensions have been proposed. \citet{cascetta1996modified} introduced the C-Logit model, which adds a commonality factor to the utility function to account for route overlap and reduce substitution bias. \citet{ben1999discrete} proposed the Path Size Logit (PSL) model, which simplifies the computation of similarity effects by using a path size factor. \citet{bovy2008factor} further developed the Path Size Correction Logit (PSCL) model to provide a more theoretically grounded correction term.
While these models improve behavioral realism by partially accounting for route similarity, they all suffer from two common limitations: their performance is highly sensitive to the construction of the choice set, and the effectiveness of the similarity correction depends heavily on how the correction factor is defined and parameterized. Beyond these explicit similarity corrections, the Mother Logit \citep{mcfadden1977application} provides a more general similarity-based structure, where utilities depend on attributes of other alternatives. Unlike models that employ explicit similarity measures, the Mother Logit captures substitution patterns through cross-effect among alternatives.

Generalized Extreme Value (GEV) models constitute a class of discrete choice models that generalize the Multinomial Logit (MNL) by allowing correlation among alternatives through a flexible error structure defined by the GEV distribution. To address the stochastic user equilibrium problem in transportation networks, \citet{chu1989paired} proposed the Paired Combinatorial Logit (PCL) and \citet{vovsha1997application} proposed the Cross-Nested Logit (CNL). PCL assumes that travelers make decisions based on pairwise comparisons of alternatives, while CNL models allocate alternatives across multiple overlapping nests to capture route similarity. Building on this line of research, \citet{bekhor2001stochastic} introduced the Generalized Nested Logit (GNL) model, in which the nesting coefficients vary across nests, offering additional flexibility in capturing substitution patterns. Although these models relax the IIA assumption and offer improved behavioral realism, they become increasingly complex as the network grows. For instance, the number of pairwise comparisons in PCL increases quadratically with the number of alternatives, resulting in high computational cost. Moreover, the prediction performance of these models remains strongly dependent on the specification of the predefined choice set.

In addition to GEV-based models, several non-GEV approaches have been developed to relax the IIA assumption. One such model is the Multinomial Probit (MNP) model proposed by \citet{daganzo1977stochastic}, which assumes that the error terms in the utility function follow a multivariate normal distribution, allowing for flexible correlation patterns among alternatives. However, the main limitation of the MNP model lies in its computational complexity, as calculating choice probabilities requires multidimensional integration.
\citet{mcfadden2000mixed} proposed the Mixed Logit model, which extends the standard logit framework by allowing model coefficients to vary randomly across individuals. This model can approximate any random utility model and accounts for both observed and unobserved heterogeneity, making it a flexible alternative to MNL.

Distinct from the above, \citet{fosgerau2013link} proposed the RL model and its extension, the Link Size Recursive Logit (LS-RL) model, which do not require explicit generation of the choice set prior to estimation. However, the link size attribute in LS-RL is still influenced by the IIA property, as will be further discussed in section 4. 
To address this issue, \citet{mai2015nested} introduced the Nested Recursive Logit (NRL) model by allowing link-specific scale parameters, thereby relaxing the IIA assumption. Later, \citet{mai2018decomposition} proposed the Mixed Recursive Logit (MRL) model, which assumes that the coefficients of the deterministic instantaneous utility are normally distributed to capture unobserved heterogeneity. 
While these extensions improve behavioral realism, they significantly increase computational complexity due to the difficulty of calculating the value function. Furthermore, they still rely on strong distributional and structural assumptions regarding scale parameters and utility specifications. Beyond distributional extensions, \citet{mai2017similarities} proposed a regret-maximization-based RL model structurally equivalent to the Mother Logit \citep{mcfadden1977application}. More recently, \citet{fosgerau2022perturbed} reformulated route choice as a network-level perturbed-utility assignment with flow conservation, enabling estimation through linear regression and prediction via convex optimization, thereby eliminating explicit choice-set enumeration. 


\subsection{Data-driven route choice models}
With the rapid increase in the availability of large-scale trajectory and sensor data, data-driven approaches have gained considerable attention in route choice modeling. These methods leverage machine learning and statistical techniques to directly learn patterns from data, often achieving significantly higher prediction accuracy than traditional theory-based models.

One of the earliest data-driven studies on route choice was conducted by \citet{yamamoto2002drivers}, who used decision trees to predict travelers' selections between two expressways. Building on this idea, \citet{lee2010hybrid} developed a logistic regression tree to model travelers’ route choices when provided with traffic information.
Random Forest (RF), an ensemble learning method that aggregates multiple decision trees trained on random subsets of features, has also been widely adopted. \citet{tribby2017analyzing} applied RF to pedestrian route choice modeling and found that it offered higher predictive accuracy than traditional models without requiring predefined utility structures. Similarly, \citet{schmid2022modeling} compared RF with Multinomial Logit (MNL), Mixed Logit, Artificial Neural Networks (ANN), and Support Vector Machines (SVM), and concluded that RF outperforms these models in terms of prediction accuracy. They also noted substantial differences in variable importance rankings across RF, MNL, and Mixed Logit.
While tree-based methods offer relatively high interpretability, they often fall short in prediction performance when compared to more advanced machine learning models. Moreover, these methods are limited in that they cannot easily accommodate the full set of possible routes in large-scale networks.

Neural Networks (NN) are a class of machine learning models capable of capturing complex, non-linear relationships in data through multi-layered processing structures. In route choice modeling, most NN-based approaches take individual and contextual features as input and output the predicted choice probabilities.
\citet{yang1993exploration} proposed one of the earliest NN-based models to predict whether a traveler would choose a freeway or a side road. \citet{dia2007modelling} further extended this approach by developing an agent-based neural network to examine how socio-economic, contextual, and informational factors influence commuter route choices.
More recent studies have demonstrated that NN-based models often outperform traditional statistical methods in terms of prediction accuracy \citep{politis2023route,lai2019understanding}, owing to their ability to automatically learn hierarchical representations from data.

Reinforcement learning has been applied to route choice models to learn the optimal policy for selecting the next link. In reinforcement learning, an agent interacts with an environment by observing a state, choosing an action, and receiving a predefined reward, with the objective of maximizing the cumulative reward. In contrast, the inverse reinforcement learning aims to recover the underlying reward function of each action by imitating experts' behavior. \cite{zhao2023deep} proposed a deep inverse reinforcement learning-based route choice model, assuming that the reward function and policy function are context-dependent. Similar to RL, the proposed model is also link-based. \cite{wei2014day} developed a theoretical day-to-day dynamic route choice model based on reinforcement learning, focusing on the evolution of travelers’ route choices over repeated days. The study is methodological rather than empirical, and it formulates the learning process without validating the model on real-world data. From a theoretical perspective, RL can be seen as a special case of inverse reinforcement learning when the environment is fully observable and the state transitions are deterministic—that is, the transition probability from one state to the next is 1 if a valid action exists.

Recurrent Neural Networks (RNNs) are well-suited for modeling sequential data due to their internal memory mechanisms, which allow them to retain information from previous inputs. As a result, RNNs have been widely used in natural language processing and time series analysis. In the context of route choice modeling, a route can be represented as a sequence of links, making RNNs a natural fit for learning sequential travel patterns.
\citet{dong2022utility} proposed a utility-based hybrid Transformer–LSTM model to analyze drivers’ route choice behavior. \citet{wang2021personalized} incorporated an RNN into an enhanced $A^*$ search algorithm to provide personalized route recommendations.
\subsection{Integration of deep learning and discrete choice models}
Despite the high predictive accuracy of deep learning methods, they are often criticized as black-box models due to the difficulty in interpreting the relationship between inputs and outputs. \citet{van2022choice} discussed the opportunity for using machine learning to enhance choice modeling ability, such as finding utility functions and capturing systematic heterogeneity. In recent years, research on integrating deep learning with discrete choice models has gained increasing attention as a way to combine the strengths of both approaches. 

\citet{sifringer2020enhancing} integrated ANNs into the MNL and NL models. In their hybrid framework, the systematic utility is decomposed into two components: a knowledge-driven part derived from a predefined utility function, and a data-driven part generated by the ANN. Estimation results demonstrate that the proposed models (L-MNL and L-NL) outperform their traditional counterparts.
Similarly, \citet{wong2018discriminative} incorporated a restricted Boltzmann machine (RBM) into the MNL model to capture latent behavioral attributes.
\citet{han2020neural} proposed the TasteNet-MNL model, which also decomposes the utility function into knowledge-driven and data-driven components. However, in this model, the data-driven component determines the parameters of a linear utility function, rather than directly outputting utility values.

{There are also several studies that incorporate GNNs into discrete choice models. \citet{cheng2026graph} demonstrated that Nested Logit and Spatially Correlated Logit models can be represented as specific GNN-based logit formulations through appropriately designed graph structures and aggregation functions. \citet{villarraga2025designing} generalized the restrictive linear network-effect component of conventional spatial discrete choice models using a GNN, enabling more flexible and nonlinear social and spatial interactions. Both studies use GNNs to replace the conventional utility specification or one of its components to capture the difference between nodes. However, these models focus on static discrete choices rather than sequential decision-making. Therefore, they cannot be directly applied to link-based route choice modeling.}

\begin{figure}
  \centering
  \includegraphics[width=300pt]{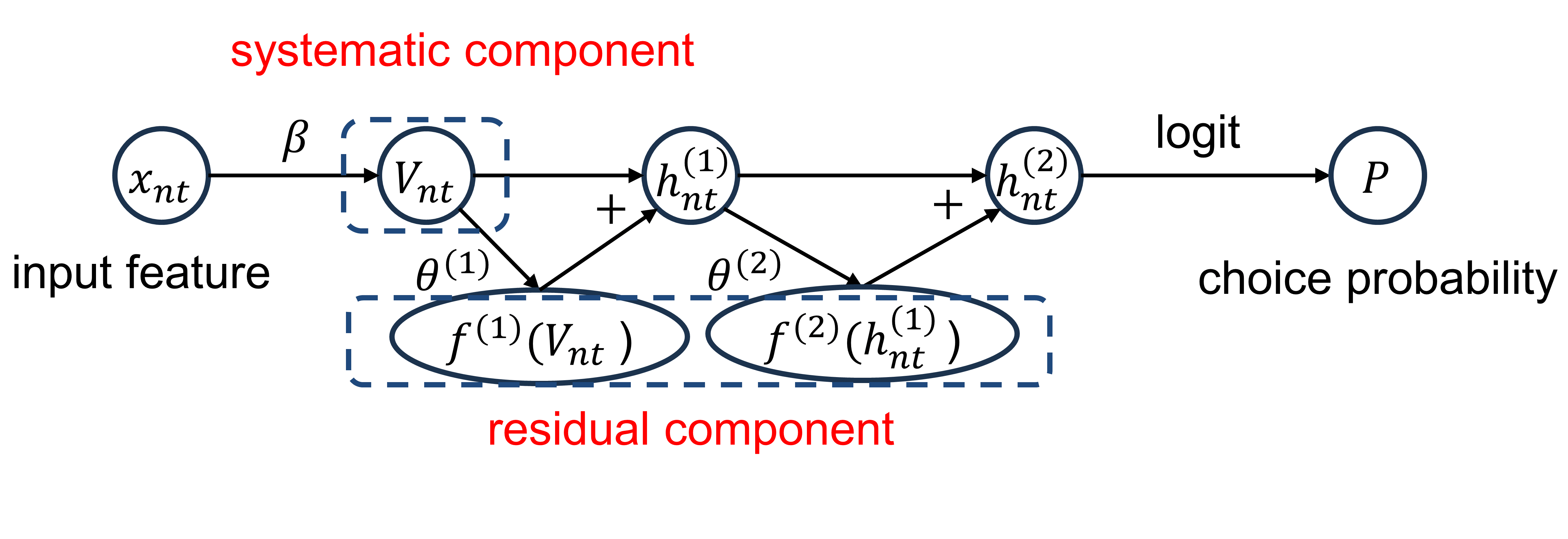}
  \caption{Framework of ResLogit.}
  \label{reslogit}
\end{figure}

The study most closely related to our work is ResLogit \citep{wong2021reslogit}. Figure \ref{reslogit} shows the framework of a two-layer ResLogit. The utility of choosing alternative $i$ from $J$ alternatives by individual $n$ in a choice task $t$ is defined as:
\begin{equation}
  U_{int}=V_{int}+g_{int}+\epsilon_{int}
\end{equation}
where $V_{int}$ and $g_{int}$ are the systematic and residual components of $U_{int}$, respectively, and $\epsilon_{int}$ is a random error term. Let $V_{nt}$ denote a $J \times 1$ vector in which $i$-th element is $V_{int}$, which is computed using a linear function. Similarly, let $g_{nt}$ be a $J \times 1$ vector whose $i$-th element is $g_{int}$. The residual component $g_{nt}$ is computed through a residual neural network as follows:

\begin{equation}
  g_{nt}=-\sum_{m=1}^M \textrm{ln}\( \mathbf{1}_{J \times 1}+\textrm{exp}\(\theta^{(m)}h_{nt}^{(m-1)}\)\)
\end{equation}
where $\mathbf{1}_{J \times 1}$ is an all-ones vector of size $J \times 1$, $h_{nt}^{(m)}$ denotes the output of the $m$-th hidden layer, and $\theta^{(m)}$ is the corresponding weight parameter.

The input layer of the network is initialized with the systematic utility component:
\begin{equation}
h_{nt}^{(0)} = V_{nt}
\end{equation}

For the $m$-th block ($m \geq 1$), the hidden representation is recursively updated as:
\begin{equation}
h_{nt}^{(m)} = h_{nt}^{(m-1)} -f(h_{nt}^{(m-1)})  
\end{equation}
\begin{equation}
f(h_{nt}^{(m-1)}) = \ln\left( \mathbf{1}_{J \times 1} + \exp\left( \theta^{(m)} h_{nt}^{(m-1)} \right) \right)
\end{equation}

The weight matrices $\theta^{(m)}$ capture the cross-effect among alternatives at each layer. A positive element in $i$-th row and $j$-th column $\theta^{(m)}_{ij}$ represent negative cross-effect of alternative $i$ on $j$. If all $\theta^{(m)}$ are identity matrices, there is no cross-effect
between alternatives, and the ResLogit model reduces to the standard MNL model.
Additionally, the use of a skip connection structure helps mitigate the problems of exploding or vanishing gradients during backpropagation, thereby improving the stability of training in deeper networks.

In path-based route choice modeling, ResLogit considers feasible paths as alternatives and uses the residual component to account for cross-effects arising from correlations among overlapping routes. However, it is challenging to determine which path attributes should be extracted. 

\section{Methodology}

In this section, the proposed hybrid route choice models are introduced. First, the Residual Recursive Logit model (Res-RL) is formulated by extending the standard RL model following the established ResLogit approach \citep{wong2021reslogit}. However, as demonstrated later, Res-RL cannot fully capture cross-effects between paths. To address this limitation, a novel model named the Residual Directed Graph Convolutional Network Recursive Logit model (ResDGCN-RL) is proposed. ResDGCN-RL constitutes the core contribution of this study. To facilitate understanding of the proposed models, notation lists for each model are provided in Appendix~A.

\subsection{Recursive Logit model and its variants}
\begin{figure}[t]
  \centering
  \includegraphics[width=\textwidth]{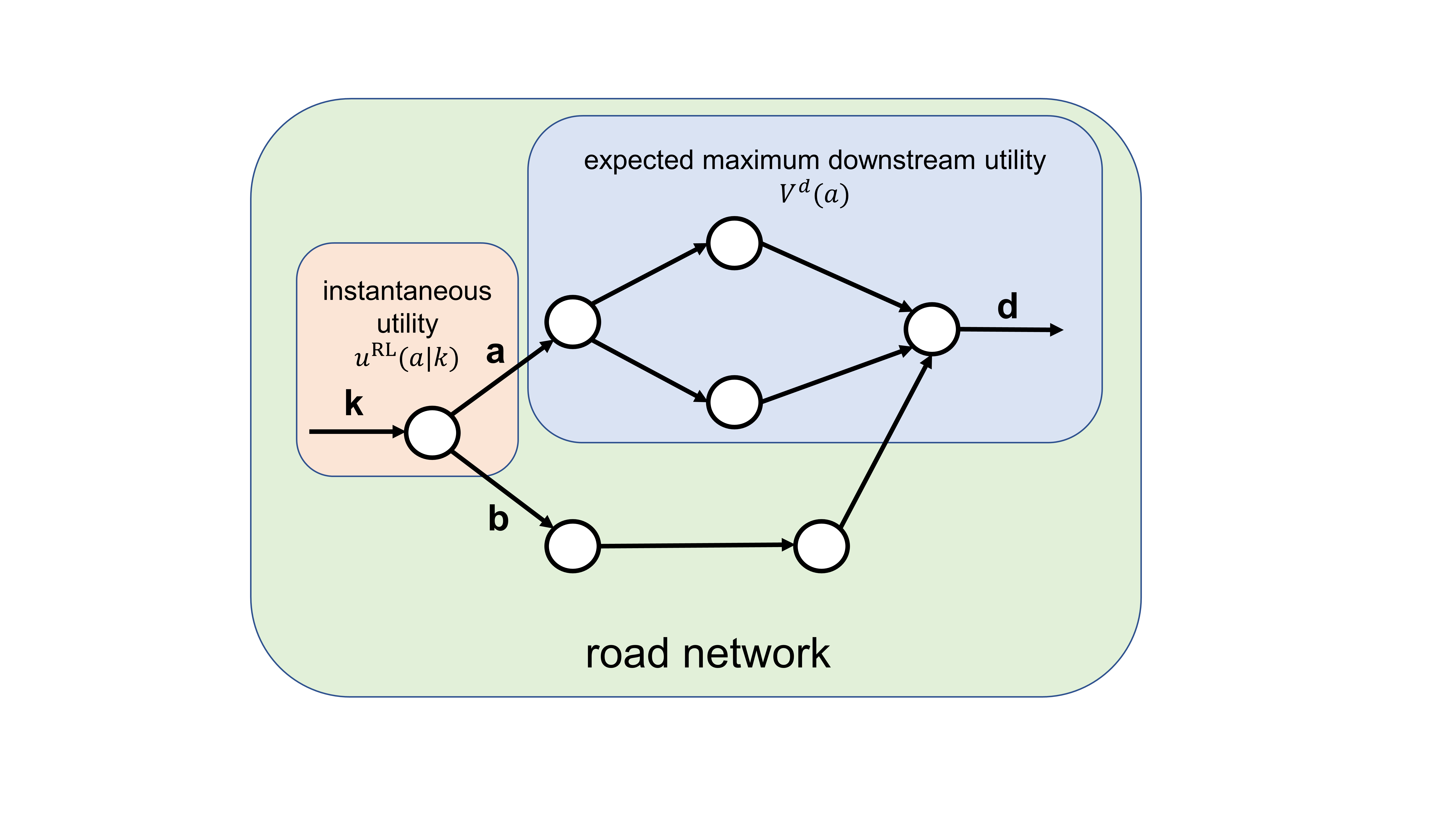}
  \caption{Illustrative road network for Recursive Logit model}
  \label{rlex}
\end{figure}

As a preparation, the existing RL formulation is summarized at first. Consider a road network $G=\langle\mathcal{A},\mathcal{V}\rangle$ where $\mathcal{A}$ and $\mathcal{V}$ are the sets of nodes and links and $A$ is the adjacency matrix of links. In RL model, route choice problem is transformed into a sequence of link choice problems. Figure \ref{rlex} showed a simple road network in which travelers, currently at link $k$ with destination $d$, are assumed to choose the outgoing link $a$ that maximizes the sum of the instantaneous utility $u^{\textrm{RL}}(a|k;\phi)$ and expected maximum downstream utility $V^d(a)$. The instantaneous utility of choosing link $a$ from $k$ can be expressed as:
\begin{equation}
  u^{\textrm{RL}}(a|k;\phi)=v^{\textrm{RL}}(a|k;\phi)+\mu\epsilon(a)
\end{equation}
where $v^{\textrm{RL}}(a|k;\phi)$ is a deterministic utility of traveling from link $k$ to $a$ usually calculated by a predefined linear function with parameters $\phi$, $\mu$ is scale parameter and assumed to be the same across all links, and $\epsilon(a)$ are i.i.d extreme type 1 error terms.

The expected maximum downstream utility at link $k$ when the destination is $d$ is calculated by Bellman's equation:
\begin{equation}
\label{bellman}
  V^d\(k\)=\mathbb{E}\left[\max_{a \in O(k)}\(u\(a|k\)+V^d\(a\)+\mu\epsilon\(a\)\)\right]
\end{equation}
where $O\(k\)$ means the set of outgoing links of link $k$ and $V^d\(d\)=0$. By applying the logsum operator, Equation \eqref{bellman} can be expressed more compactly as follows:
\begin{equation}
V^d(k) = 
\begin{cases}
\mu \ln \sum_{a \in \mathcal{V}} \delta(a|k)\, e^{\tfrac{1}{\mu}\left( v^{\textrm{RL}}(a|k) + V^d(a) \right)}, & \forall k \in \mathcal{V} \\[1.2ex]
0, & k = d
\end{cases}
\end{equation}

The probability of choosing link $a$ from current link $k$ with destination $d$ follows an MNL model.
\begin{equation}
  P^d\(a|k\)=\delta\(a|k\)\frac{e^{\frac{1}{\mu}\(v^{\textrm{RL}}\(a|k\)+V^d\(a\)\)}}{\sum_{a'\in A\(k\)}e^{\frac{1}{\mu}\(v^{\textrm{RL}}\(a'|k\)+V^d\(a'\)\)}}=\delta\(a|k\)e^{\frac{1}{\mu}\(v^{\textrm{RL}}\(a|k\)+V^d\(a\)-V^d\(k\)\)}
\end{equation}
where $\delta\(a|k\)$ is a dummy variable, equals one if $A_{ka}=1$.

The path utility in the RL model can be computed as the sum of the utilities of all actions along the path. Consider a path
$\sigma=(a_0,a_1,\ldots,a_d)$, where $a_i$ denote the $i$-th link of the trajectory. The path choice probability $P(\sigma)$ can be calculated as:

\begin{equation}
P(\sigma)=e^{v^{\textrm{RL}}(a_1|a_0)+V^{a_d}(a_1)-V^{a_d}(a_0)+v^{\textrm{RL}}(a_2|a_1)+V^{a_d}(a_2)-V^{a_d}(a_1)+\cdots+v^{\textrm{RL}}(a_d|a_{d-1})+V^{a_d}(a_d)-V^{a_d}(a_{d-1})}
\end{equation}
and the above equation can be arranged as follows:
\begin{equation}
P(\sigma)=\frac{e^{v^{\textrm{RL}}(a_1|a_0)+v^{\textrm{RL}}(a_2|a_1)+\cdots+v^{\textrm{RL}}(a_d|a_{d-1})}}{e^{V^{a_d}(a_0)}}
\end{equation}

Thus, Within the RL model, for a given OD pair, the probability of choosing a particular trajectory increases exponentially with the sum of the action utilities along that trajectory. If the network is acyclic so that the set of feasible trajectories is finite, then RL behaves the same as an MNL model that enumerates all feasible paths. Moreover, the ratio of the choice probabilities of two trajectories equals the exponential of the difference between their total action utilities (i.e., the sum of action utilities along each trajectory), which implies the IIA property.

In order to overcome correlation in path utilities due to physical overlapping,  \citet{fosgerau2013link} propose LS-RL, where the link-size attribute is defined as the OD-specific expected link flow. This follows a similar rationale to the Path Size logit model in addressing route overlap \citep{ben1999discrete}.  Consider a destination d, $G^a$ is a vector of size $\lvert \mathcal{V} \rvert$ with the $a$-th element as 1 and all other elements as 0 and the destination is $d$, $\lvert \mathcal{V} \rvert$ is the number of links. The choice probability matrix $P^d$, with elements $P^d_{ka} = P^d(a|k)$, is computed using a standard RL model with predefined parameters.. The expected link flow for an origin–destination pair $(o,d)$, denoted as $F^{od}$, can be obtained by solving the following linear system:
\begin{equation}
\label{eq:linkflow}
  \(I-\(P^d\)^T\)F^{od}=G^{o}
\end{equation}

 By incorporating the expected link flow as a link size attribute into the utility function, the LS-RL model can mitigate the IIA problem caused by overlapping routes. However, this approach introduces substantial complexity, as it requires the utility parameters to be predefined and the link size attribute to be recalculated separately for each origin–destination (OD) pair. That is, when predicting trajectories for different OD pairs, an OD-specific expected link flow must be computed each time.

NRL \citep{mai2015nested} can also solve the physical overlapping problem. The difference between NRL and
RL is that the scale parameter in NRL is not a constant but depends on the characteristics of link. Unlike RL, whose value function can be obtained by solving a linear system, NRL requires an iterative procedure to approximate the solution due to the varying scale parameter across links.

\subsection{Residual Recursive Logit model}

Following the ResLogit structure \citep{wong2021reslogit}, RL is extended to a hybrid link-based route choice model by incorporating a neural network and term the resulting model Res-RL. This extension relaxes the IIA property of RL and improves prediction accuracy. However, as we will see later, this Res-RL does not preserve an advantage of ResLogit, namely, ability to capture full cross-effects between paths.

\subsubsection{Formulation of Res-RL}
The instantaneous utility of choosing link $a$ from $k$ is defined as:
\begin{equation}
  u^r(a|k)=v^r(a|k)+g^r(a|k)+\mu\epsilon(a)
\end{equation}
Here, $v^r(a|k)$ and $g^r(a|k)$ denote the systematic and residual components, respectively, for choosing link $a$ from node $k$ in the Res-RL model. The random term $\epsilon(a)$ is assumed to be i.i.d. extreme value type I, and $\mu$ represents the scale parameter, which is set to $\mu=1$ for simplicity. The systematic component is specified as a linear function of the action features.

The residual component is calculated by:
\begin{equation}
  G^r=-\sum_{m=1}^M \textrm{ln}\(\frac{1}{2}\(\mathbf{1}_{\lvert V \rvert \times \lvert V \rvert}+\textrm{exp}\(h^{r,m-1}\theta^{r,m}\)\)\)\odot A
\end{equation}
where $G^r$ is a $|\mathcal{V}| \times |\mathcal{V}|$ matrix whose $(k,a)$-th element represents the residual utility $g^r(a|k)$ of choosing link $a$ from link $k$. Here, $h^{r,m-1}$ is the input to the $m$-th residual layer, and $\theta^{r,m}$ is the corresponding weight matrix of size $|\mathcal{V}| \times |\mathcal{V}|$. The operator $\ln(\cdot)$ and exponential are applied element-wise, and $\odot A$ masks out invalid transitions. The inclusion of the $\frac{1}{2}(\cdot)$ term ensures that when the weight matrices are zero, the residual component $g^r(a|k)$ is also zero, allowing the model to revert to the standard RL formulation.

The hidden layers are recursively defined as follows. The input layer is initialized using the systematic component:
\begin{equation}
  h^{r,0}_{ka}=v^r(a|k)
\end{equation}

Then, for each residual layer $m = 1, 2, \dots, M$, the hidden representation is updated via:
\begin{equation}
  h^{r,m}=h^{r,m-1}-\textrm{ln}\(\frac{1}{2}\(\mathbf{1}_{\lvert \mathcal{V} \rvert \times \lvert \mathcal{V} \rvert}+\textrm{exp}\(h^{r,m-1}\theta^{r,m}\)\)\)\odot A \quad \forall m\geq1
\end{equation}

Therefore, the output layer is the deterministic component of instantaneous utility:
\begin{equation}
    h^{r,M}_{ka}=v^r(a|k)+g^r(a|k)
\end{equation}

\subsubsection{Property of Res-RL}
One of the most important properties of Res-RL, inherited from ResLogit, is the interpretability of the weight matrix, which captures the interaction effects between travel actions. To better illustrate this property, the \textit{cross-effect}, \textit{outgoing link pair}, and \textit{intersection cross-effect} are defined as follows:

\begin{description}
  \item[Cross-effect] {The dependence of the utility of one action on the utility of another action.}
  \item[Outgoing link pair]  The outgoing link pair consists of links that share a common node as their origins.
\[
\{\,((v,u),(v,w)) \mid (v,u)\in \mathcal{V}, (v,w)\in \mathcal{V}, u \neq w \,\}
\]
  \item[Intersection cross-effect] {The \textit{intersection cross-effect} describes how the utility of one action depends on that of another action when both actions originate from the same link and lead to links forming an outgoing link pair.}
\end{description}

The element $\theta^{r,m}_{a'a}$ captures the \textit{\textit{intersection cross-effect}}, representing how the utility of traveling to link $a'$ influences that of traveling to link $a$. In a simple one-layer Res-RL, the utility of traveling from link $k$ to link $a$ is defined as:
\begin{equation}
  u^r(a|k)=v^r(a|k)-
  \ln\(\frac{1}{2}\(1+\exp\(\sum_{a' \in O(k)} v^r(a'|k)\times \theta^{r,1}_{a'a}\)\)\)
  +\epsilon(a)
\end{equation}

The term $\exp\!\left(\sum_{a' \in O(k)} v^r(a'|k)\,\theta^{r,1}_{a'a}\right)$ represents how the utilities of choosing other links influence the utility of choosing link $a$. To explicitly quantify this cross-effect, the partial derivative of $u^r(a|k)$ with respect to $v^r(a'|k)$ for $a, a' \in A(k)$ and $a \neq a'$ is computed as:
\begin{equation}
  \frac{\partial u^r(a|k)}{\partial v^r(a'|k)}
  = \frac{1-2\exp\!\left(-g^{r}(a|k)\right)}{2\exp \(-g^{r}(a|k)\)}\,
    \theta^{r,1}_{a'a}
\end{equation}
It is easy to prove that $\frac{1-2\exp\!\left(-g^{r}(a|k)\right)}{2\exp \(-g^{r}(a|k)\)}$ is smaller than 0 because $g^{r}(a|k)$ is smaller than $\ln 2$. Accordingly, a positive value of $\theta^{r,1}_{a'a}$ indicates that a higher $v^r(a'|k)$ leads to a lower $u^r(a|k)$, and vice versa.


It should be noted that Res-RL does not satisfy the IIA property.
 Consider two paths
$\sigma_a=(a_1,a_2,\dots,a_{|a|})$ and $\sigma_b=(b_1,b_2,\dots,b_{|b|})$
connecting the same OD pair, where $a_i$ and $b_i$ denote links, and $|a|$ and $|b|$ denote the lengths of
$\sigma_a$ and $\sigma_b$, respectively. The ratio of their choice probabilities is given by
\begin{equation}
\label{eq:resrl_ratio}
\frac{P(\sigma_a)}{P(\sigma_b)}
=
\exp\!\Bigg(
\sum_{i=2}^{|a|}\big(v^r(a_i|a_{i-1})+g^r(a_i|a_{i-1})\big)
-
\sum_{i=2}^{|b|}\big(v^r(b_i|b_{i-1})+g^r(b_i|b_{i-1})\big)
\Bigg)
\end{equation}
Equation~\eqref{eq:resrl_ratio} demonstrates that the ratio of choice probabilities of two paths is determined by the differences in their total systematic utilities and residual utilities. As derived above, for each $i\in\{2,\dots,|a|\}$, the residual term $g^r(a_i|a_{i-1})$
is computed based on $\{v^r(a'|a_{i-1}): a'\in O(a_{i-1})\}$, and similarly for $\sigma_b$.
Therefore, $\frac{P(\sigma_a)}{P(\sigma_b)}$ depends on alternatives beyond $\sigma_a$
and $\sigma_b$, implying that the IIA property is relaxed in Res-RL.

However, although Res-RL relaxes the IIA property, it may still fail
to fully capture the cross-effects between paths. In equation \eqref{eq:resrl_ratio}, $\frac{P(\sigma_a)}{P(\sigma_b)}$ depends only on
$S_a=\{v^r(a'|a_{i-1}): a'\in O(a_{i-1}),\, 2\leq i \leq |a|\}$ and
$S_b=\{v^r(b'|b_{i-1}): b'\in O(b_{i-1}),\, 2\leq i \leq |b|\}$.
Consequently, if the systematic utility of any action outside $S_a$ and $S_b$ changes,
or if such an action becomes unavailable, $\frac{P(\sigma_a)}{P(\sigma_b)}$ remains
unchanged, which can lead to missing substitution pattern.  An illustrative example is provided in Section \ref{ill_app}.

\subsection{Residual Directed Graph Convolutional Network Recursive Logit model}
To better capture the cross-effect between paths, ResDGCN-RL is proposed, which extends Res-RL by integrating directed graph convolutional layers into the residual utility formulation. {The model structure is similar to that of \citet{cheng2026graph}, as both models use GNN-based message passing to propagate utility and incorporate the resulting output as a residual component.} Leveraging the message-passing mechanism of GNNs, a broader set of actions can influence the choice probability ratio between two paths.

\subsubsection{Directed Graph Convolutional Network (DGCN)}
\label{dgcnn}
Graph Convolutional Network (GCN) is widely used to process data with graph structures due to its ability to leverage both the feature of nodes and spatial relationships \citep{scarselli2008graph}. However, traditional GCN assumes that the graph is undirected, which contradicts the nature of road networks, where links have distinct origins and destinations. \cite{tong2020directed} proposed a Directed Graph Convolutional Network (DGCN) to take the directed graph feature into consideration by first-degree and second-order proximity.

First, GCN is introduced to establish key notations relevant to DGCN. The goal of GCN is to learn new features from the graph $G=\langle\mathcal{A},\mathcal{V}\rangle$ with adjacency matrix $A$. The input of GCN is an $N\times D$ matrix $X$, where $N$ is the number of nodes and $D$ is the number of features. The output of GCN is an $N\times F$ matrix $Z$, where $F$ is the dimension of learned representations. 

For each hidden layer $h^{(m)}$:
\begin{equation}
  h^{(m)}=\sigma\(\widetilde{D}^{-\frac{1}{2}}\widetilde{A}\widetilde{D}^{-\frac{1}{2}}h^{(m-1)}W^{(m)}\) \quad \forall m\geq1
\end{equation}
where $\sigma$ is the activation function, the term $\widetilde{A}=A+I$ where $I$ is identity matrix to enforce self-loops and $\widetilde{D}$ is the degree matrix of $\widetilde{A}$. $W^{(m)}$ is a learnable weight matrix of $m$-th hidden layer.

In DGCN, different types of proximities are represented by distinct adjacency matrices.

The first-order proximity is defined as
\begin{equation}
  A_F(i,j) = (A^{\mathrm{sym}})_{ij},
\end{equation}
where $A^{\mathrm{sym}}$ is the symmetrized adjacency matrix given by
\begin{equation}
  (A^{\mathrm{sym}})_{ij} = \mathbf{1}\{A_{ij}+A_{ji}>0\},
\end{equation}
with $\mathbf{1}\{\cdot\}$ denoting the indicator function, which equals $1$ if the condition holds and $0$ otherwise.

The second-order proximities are defined as
\begin{equation}
  A_{S_{\mathrm{in}}}(i,j) = \sum_k \frac{A_{i,k}A_{j,k}}{\sum_v A_{v,k}},
\end{equation}
\begin{equation}
  A_{S_{\mathrm{out}}}(i,j) = \sum_k \frac{A_{k,i}A_{k,j}}{\sum_v A_{k,v}},
\end{equation}
where $A_{S_{\mathrm{in}}}$ captures common successors of nodes $i$ and $j$ 
(in-degree proximity), whereas $A_{S_{\mathrm{out}}}$ captures their common predecessors 
(out-degree proximity). 

Based on these adjacency matrices, the corresponding normalized adajacency matrices are constructed as:
\begin{equation}
\label{z1}
  Z_F = \widetilde{D}_F^{-\frac{1}{2}}\widetilde{A}_F\widetilde{D}_F^{-\frac{1}{2}},
\end{equation}
\begin{equation}
\label{z2}
  Z_{S_{\textrm{in}}} = \widetilde{D}_{S_{\textrm{in}}}^{-\frac{1}{2}}
  \widetilde{A}_{S_{\textrm{in}}}
  \widetilde{D}_{S_{\textrm{in}}}^{-\frac{1}{2}},
\end{equation}
\begin{equation}
\label{z3}
  Z_{S_{\textrm{out}}} = \widetilde{D}_{S_{\textrm{out}}}^{-\frac{1}{2}}
  \widetilde{A}_{S_{\textrm{out}}}
  \widetilde{D}_{S_{\textrm{out}}}^{-\frac{1}{2}},
\end{equation}
where $\widetilde{D}_F$, $\widetilde{D}_{S_{\textrm{in}}}$, and $\widetilde{D}_{S_{\textrm{out}}}$ 
are the degree matrices including self-loops, consistent with the GCN formulation.

Finally, the hidden representation at layer $m$ is updated as:
\begin{equation}
  h^{(m)} = \textrm{softmax}\!\left(
    \textrm{Concat}\!\left(
      Z_F h^{(m-1)} W_1^{(m)},
      \, \delta_1 Z_{S_{\textrm{in}}} h^{(m-1)} W_1^{(m)},
      \, \delta_2 Z_{S_{\textrm{out}}} h^{(m-1)} W_1^{(m)}
    \right)\right), \quad \forall m \geq 1,
\end{equation}
where $\textrm{Concat}(\cdot)$ denotes matrix concatenation, 
$\delta_1$ and $\delta_2$ are learnable parameters controlling the relative importance of different proximities, 
and $\textrm{softmax}$ is an activation function ensuring that each output vector forms a probability distribution.

\subsubsection{Formulation of ResDGCN-RL}
\begin{figure}[t]
  \centering
  \includegraphics[width=\textwidth]{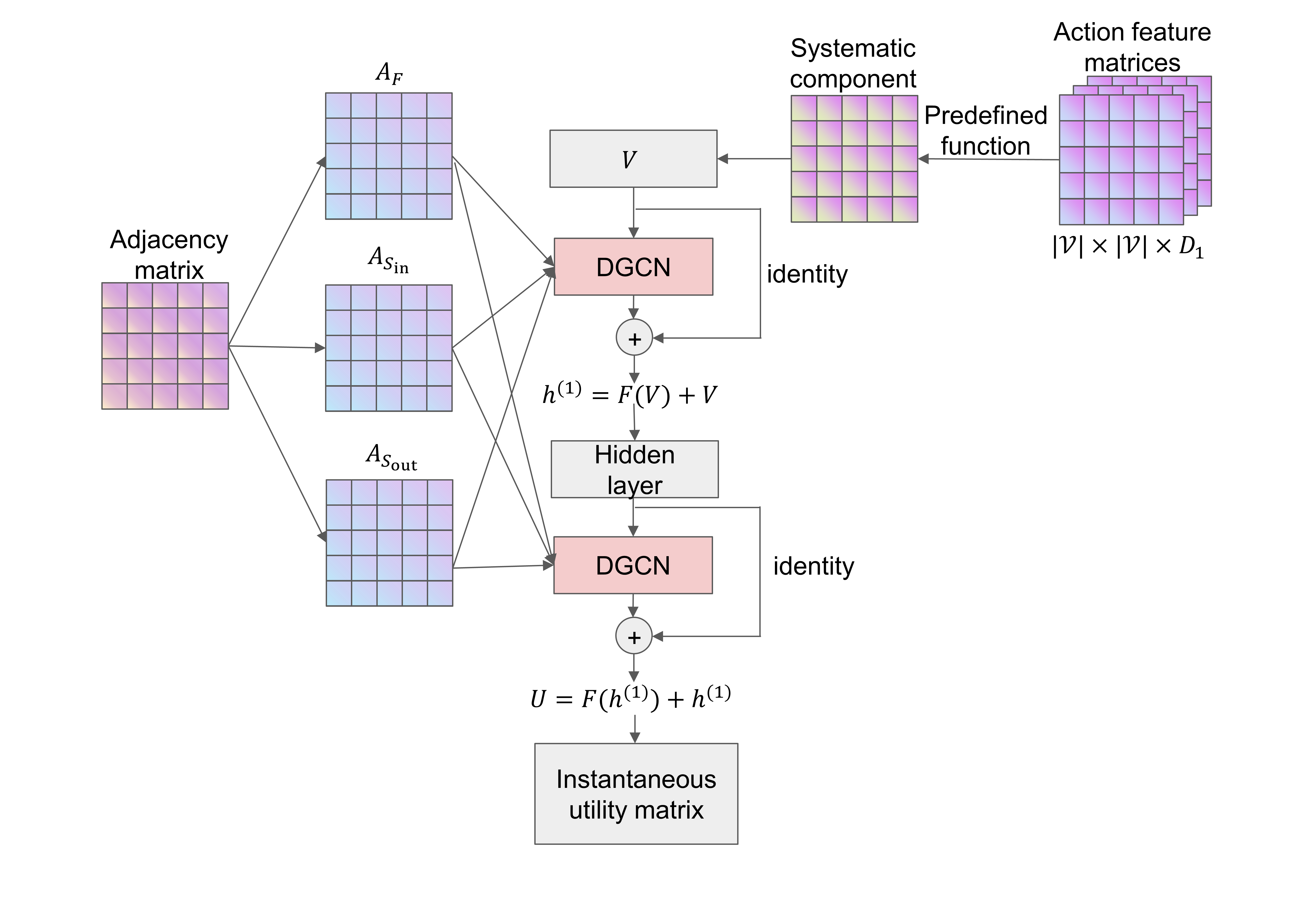}
  \caption{Framework of ResDGCN-RL. $F(\cdot)$ represents the modified DGCN propagation rule.}
  \label{resdgcnrl-fra}
\end{figure}

Inspired by DGCN, a novel model named ResDGCN-RL is proposed. The framework of ResDGCN-RL is shown in Figure \ref{resdgcnrl-fra}. The proposed model processes adjacency and systematic component matrices through DGCN layers with residual connections, generating the instantaneous utility matrix.

Specifically, The instantaneous utility of traveling from link $k$ to $a$ of ResDGCN-RL, denoted as $u^d(a|k)$, follows the same formulation as Res-RL: 
\begin{equation}
  u^d(a|k)=v^d(a|k)+g^d(a|k)+\mu\epsilon(a)
\end{equation}
where $v^d(a|k)$ and $g^d(a|k)$ are the systematic component and the residual component of $u^d(a|k)$, respectively. Similar to Res-RL, $\mu$ is set to 1.

The residual component matrix $G^d$ with elements $G^d(k,a)=g^d(a|k)$ is learned via a modified DGCN to capture link cross-effect:

\begin{equation}
  G^d=-\sum_{m=1}^M\textrm{ln}\(\frac{1}{2}\(\mathbf{1}_{\lvert A \rvert \times \lvert A \rvert}+\exp\(\(\alpha Z_F+\beta Z_{S_{\textrm{in}}}+\gamma Z_{S_{\textrm{out}}}\)h^{d,m-1}\theta^{d,m}\)\)\)\odot A
  \label{gd1}
\end{equation}

For the input layer,
\begin{equation}
    h^{d,0}_{ka}=v^{d}(a|k)
\end{equation}

For the $m$-th hidden layer $h^{d,m}$:

\begin{equation}
  h^{d,m}=h^{d,m-1}+F(h^{d,m-1})
  \label{gd2}
\end{equation}

\begin{equation}
  F(h^{d,m-1})=-\textrm{ln}\(\frac{1}{2}\(\mathbf{1}_{\lvert A \rvert \times \lvert A \rvert}+\exp\(\(\alpha Z_F+\beta Z_{S_{\textrm{in}}}+\gamma Z_{S_{\textrm{out}}}\)h^{d,m-1}\theta^{d,m}\)\)\)\odot A
  \label{gd3}
\end{equation}

where $\theta^{d,m}$ is the weight matrix in the $m$-th hidden layer, and $\alpha$, $\beta$, and $\gamma$ are learnable parameters that explicitly weight the contributions of $Z_F$, $Z_{S_{\textrm{in}}}$, and $Z_{S_{\textrm{out}}}$, respectively. The definitions of $Z_F$, $Z_{S_{\textrm{in}}}$, and $Z_{S_{\textrm{out}}}$ are the same as those in the original DGCN model, except that the proximity relations are defined over links rather than nodes.

Therefore, the output layer is the deterministic component of instantaneous utility:
\begin{equation}
    h^{d,M}_{ka}=v^d(a|k)+g^d(a|k)
\end{equation}

\subsubsection{Property of ResDGCN-RL}
\label{resd_property}
Similar to Res-RL, ResDGCN-RL also yields an interpretable weight matrix. The key difference is that Res-RL can only capture \emph{\textit{intersection cross-effect}}, whereas ResDGCN-RL extends this capability to incorporate multiple types of cross-effect patterns across the network. Moreover, the learnable parameters $\alpha$, $\beta$, and $\gamma$ explicitly quantify the relative importance of first-order and second-order proximities, thereby offering a richer behavioral interpretation.

\begin{figure}[htbp]
    \centering
    \includegraphics[width=0.8\linewidth]{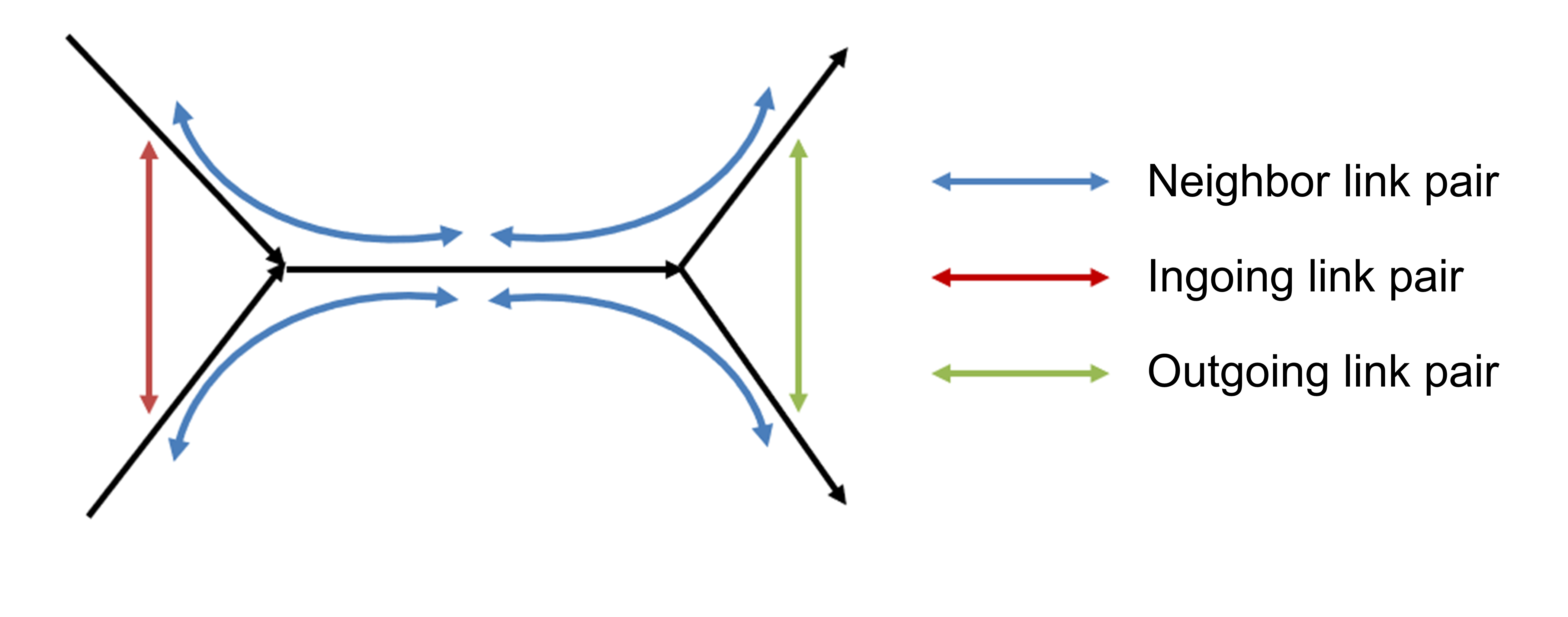}
    \caption{llustration of neighbor, ingoing and outgoing link pair.}
    \label{f1}
\end{figure}

\begin{figure}[htbp]
    \centering
    \includegraphics[width=0.5\linewidth]{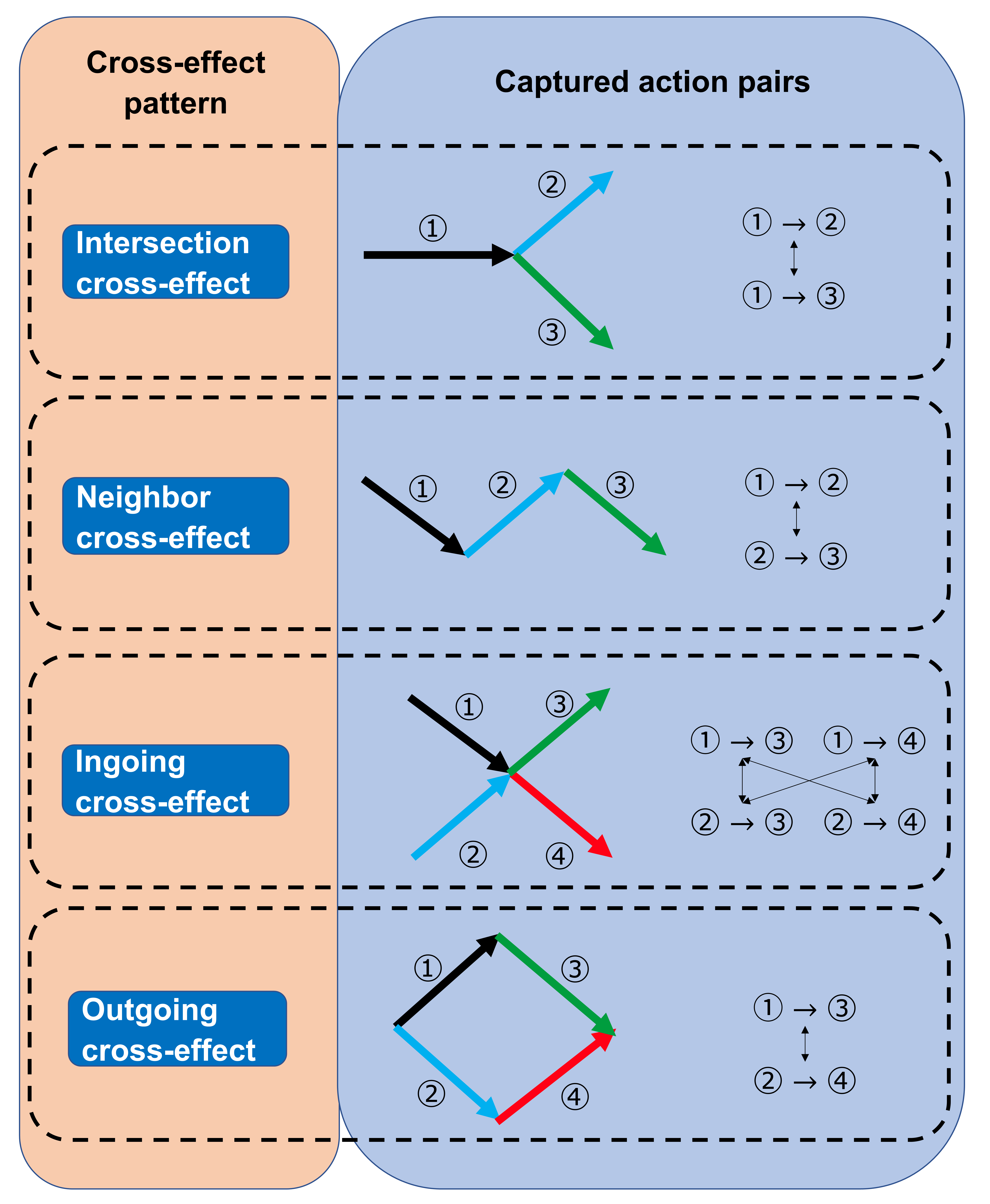}
    \caption{Illustration of four cross-effect patterns. For each pattern, the colored arrows denote links, and the diagrams on the right illustrate the action pairs whose cross-effects can be captured.}
    \label{fig:cf}
\end{figure}

To facilitate understanding of the model properties, the neighbor link pair, ingoing link pair, and other cross-effect patterns are first defined.  Figure \ref{f1} illustrates \textit{neighbor link pair}, \textit{outgoing link pair}, and \textit{ingoing link pair}, where the cross-effect between links originate from links in these link pairs can be captured by first-order and second-order proximity matrices. In most cases, two links in a road network can form only one type of link pair as defined above. Figure \ref{fig:cf} illustrates four distinct cross-effect patterns and what kind of action pairs they can capture. Notably, the link cross-effect are invariant across origin–destination pairs.

\begin{description}
  \item[Neighbor link pair] 
  A neighbor link pair consists of two directed links such that 
  the destination node of one link coincides with the origin node of the other. 
\[
\{\,((u,v),(v,w)) \mid (u,v)\in\mathcal{V}, (v,w)\in\mathcal{V}\}
\]
  \item[Ingoing link pair] 
 An ingoing link pair consists of two links that share a common node as their destinations.
\[
\{\,\{(u,v),(w,v)\} \mid (u,v)\in\mathcal{V}, (w,v)\in\mathcal{V}\}
\]
  \item[Neighbor cross-effect] The neighbor cross-effect refers to cross-effect between two actions that travel originated from links in a neighbor link pair.
  \item[Ingoing cross-effect] The \textit{ingoing cross-effect} refers to cross-effect between two actions that travel originated from links in a ingoing link pair.
  \item[Outgoing cross-effect] The \textit{outgoing cross-effect} refers to cross-effect between two actions that travel originated from links in a outgoing link pair.
\end{description}



Consider a ResDGCN-RL with only one hidden layer,
\begin{equation}
\begin{aligned}
  u^d(a|k) =\;& v^d(a|k) \\
  & - \ln\(\frac{1}{2}\(1+\exp\!\(
      \sum_{j}\sum_{i} 
        \bigl(\alpha Z_F(k,j) + \beta Z_{S_{\mathrm{in}}}(k,j) + \gamma Z_{S_{\mathrm{out}}}(k,j)\bigr)
        v^d(i\mid j)\,\theta^{d,0}_{ia}
    \)\)\)
    + \epsilon(a).
\end{aligned}
\end{equation}

The partial derivative of $u^d(a|k)$ with respect to $v^d(a'|k')$ and $a \neq a'$ is:
\begin{equation}
  \frac{\partial u^d(a|k) }{\partial v^d(a'|k')}=\frac{1-2\exp\!\left(-g^{d}(a|k)\right)}{\exp\(-g^{r}(a|k)\)}\,(\alpha Z_F(k,k')+\beta Z_{\mathrm{in}}(k,k')+\gamma Z_{\mathrm{out}}(k,k'))\theta^{d,0}_{a'a}
\end{equation}

Based on the above derivations, it can be concluded that $\theta^{d,\cdot}_{aa'}$ indicates the cross-effect of the utility of action travel to links $a'$ on the action travel to $a$, while the proximity between the origins of the two actions, together with the coefficients $\alpha$, $\beta$, and $\gamma$, can be interpreted as adjustment factors for this cross-effect.  If $k=k'$, $\theta^{d,0}_{a'a}$ captures the \textit{intersection cross-effect}, which is the only cross-effect pattern that Res-RL can capture. If $k$ and $k'$ form a neighbor link pair, $\theta^{d,0}_{a'a}$ captures the \textit{neighbor cross-effect}. The same logic applies to the other cross-effect patterns. The importance of $\alpha$, $\beta$, and $\gamma$ is further illustrated with an example in Appendix B.

{\color{red}The relationship between ResDGCN-RL and Res-RL is clarified next. In Res-RL, the $(i,j)$-th entry of the weight matrix can be nonzero only when links $i$ and $j$ form an outgoing link pair. Consequently, Res-RL can capture only \textit{intersection cross-effects}. In ResDGCN-RL, graph-based message passing allows weight matrix entries to be nonzero even when the corresponding links do not form such a pair, enabling the model to capture additional cross-effects. If the nonzero entries in ResDGCN-RL are restricted to outgoing link pairs, its structure becomes identical to that of Res-RL. Thus, Res-RL can be viewed as a special case of ResDGCN-RL. The additional parameters in ResDGCN-RL therefore extend the types of cross-effects the model can represent, rather than merely increasing its flexibility in modeling the same intersection cross-effects.}

Building on Equation~\eqref{eq:resrl_ratio}, the IIA property is also relaxed in ResDGCN-RL. Compared with Res-RL, ResDGCN-RL can incorporate richer cross-effect patterns, so the set of actions that may influence the choice probability ratio between two paths is broader. Theoretically, due to the message-passing mechanism of GNNs, increasing the number of hidden layers enlarges the spatial range over which action utilities can propagate; consequently, actions farther away in the network can affect the choice-probability ratio between two paths, thereby cross-effects between paths can be much more effectively captured, enabling more general substitution patterns to be represented.

The computational complexity of Res-RL and ResDGCN-RL is $\mathcal{O}(M\lvert \mathcal{V} \rvert^{3})$, and the space complexity is $\mathcal{O}(M\,\lvert \mathcal{V} \rvert^{2})$.

\subsection{Loss function}
In RL, the estimation method used is maximum log-likelihood estimation. The log-likelihood is defined as:
\begin{equation}
  \textrm{LL}=\sum_{n=1}^N\sum_{l=1}^{l_n-1}\textrm{ln}P^{a_{l_n}^n}(a^n_{l+1}|a^n_l)
\end{equation}
where $\textrm{LL}$ represents log-likelihood, $N$ is the number of observed trajectories, $l_n$ denotes the number of links in $n$-th trajectory and $a^n_{l}$ refers to the $l$-th link in $n$-th trajectory. 

{
To balance the trade-off between interpretability and prediction accuracy, a penalty coefficient $\lambda$ is introduced into the loss function inspired by L2 regularization. The penalty term of a model is defined by:
\begin{equation}
  \textrm{Penalty}=\sum_{m=1}^{M} \left\|\theta^{m}\right\|^2_2
\end{equation}
where $\theta^{m}$ refers to the weight matrix of the $m$-th layer, corresponding to $\theta^{r,m}$ in Res-RL or $\theta^{d,m}$ in ResDGCN-RL. Here, $\left\|\cdot\right\|_2^2$ denotes the squared Euclidean norm.

A lower penalty value indicates that the instantaneous utility relies more on the systematic component and less on the residual component. This suggests that parameters within the systematic component carry more meaningful information, enabling better interpretation of feature importance (e.g., the value of time). Since standard RL does not incorporate a deep learning component, its penalty term is considered zero.

The loss function of Res-RL and ResDGCN-RL is formulated as:
\begin{equation}
  \textrm{Loss}=-\textrm{LL}+\lambda \times \textrm{Penalty}
\end{equation}
The corresponding gradient formulations are presented in Appendix~C.

A higher $\lambda$ imposes a stronger penalty on the residual component, thereby increasing the relative share of the systematic utility in the total utility and enhancing interpretability, potentially at the cost of prediction accuracy. Conversely, a lower $\lambda$ allows the residual component to contribute more strongly, prioritizing prediction performance over interpretability. The appropriate value of $\lambda$ depends on the feature scale and the specific objective of the application. For instance, a higher $\lambda$ may be selected when understanding travel behavior for urban planning is prioritized, whereas $\lambda=0$ or a small penalty coefficient may be more appropriate for applications emphasizing prediction accuracy, such as traffic forecasting or route recommendation. Because the effect of $\lambda$ also depends on the network and trajectory distribution, it should be retuned for each dataset rather than transferred directly across applications. For a model estimated with a given $\lambda$, it is possible to quantitatively evaluate how much of its explanatory power comes from the interpretable systematic utility function and how much comes from the data-driven machine learning component. Therefore, the analyst can adjust $\lambda$ according to their objective based on these values.

For consistency in notation, throughout this paper, Res-RL  $(M = m, \lambda = l)$ is used to denote a Res-RL model with $m$ hidden layers and a penalty coefficient of $l$. The same notation applies to ResDGCN-RL.
}

\section{Mechanism demonstration}
\label{ill_app}
\begin{figure}[htbp]
    \centering
    \includegraphics[width=1.0\linewidth]{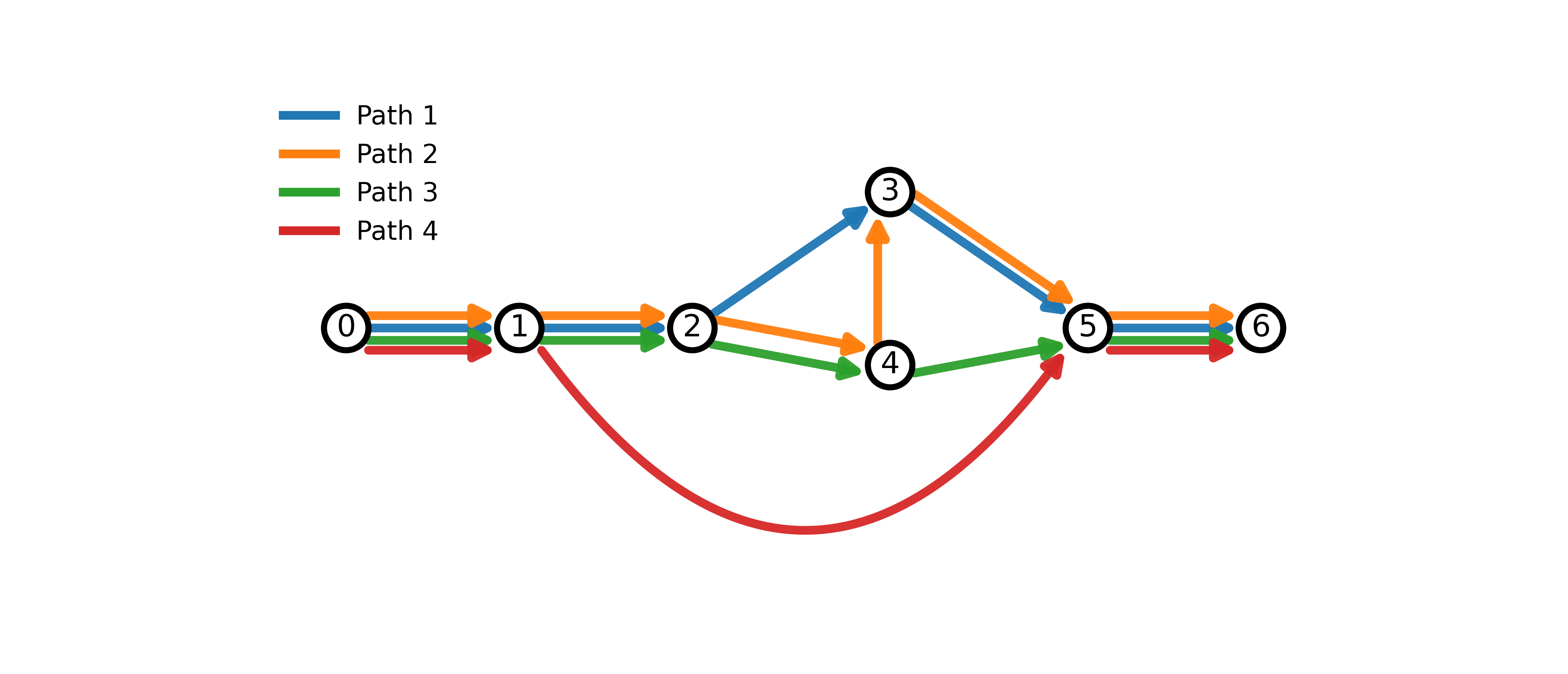}
    \caption{A simple network for illustrative example.}
    \label{fig:substitution effect}
\end{figure}

{In this section, the toy network shown in Figure~\ref{fig:substitution effect} is used to examine whether the models can learn a prespecified change in route-choice behavior from synthetic data. The assumed behavioral change represents a substitution effect that Res-RL cannot capture.
} The network consists of 7 nodes and provides four alternative paths from node 0 to node 6, which are
$0\rightarrow 1\rightarrow 2\rightarrow 3\rightarrow 5\rightarrow 6$,
$0\rightarrow 2\rightarrow 4\rightarrow 3\rightarrow 5\rightarrow 6$,
$0\rightarrow 2\rightarrow 4\rightarrow 5\rightarrow 6$, and
$0\rightarrow 1\rightarrow 5\rightarrow 6$.
These paths are labeled as Path~1, Path~2, Path~3, and Path~4, respectively.
Throughout this scenario, the only feature considered for each link from $i$ to $j$ is its travel time $t_{ij}$.
The synthetic travel-time settings and choice datasets are constructed, as shown in Tables~\ref{c4sc1} and~\ref{c4sc1data}.
To illustrate the limitation of Res-RL, the training dataset contains two periods: one for the complete network and the other for the network with removal of link $4\to 5$.
The data used in the period before removal are set to match the prediction results of RL-LS in \citet{fosgerau2013link}.
{After removing link $4\to 5$, the choice probability of Path~3 is reallocated more to Path~1 and Path~2.}


\begin{table}[htbp]
  \centering
  \caption{Link features for illustrative example.}
  \label{c4sc1}
    
  \fontsize{8pt}{10pt}\selectfont
  \begin{tabular}{c c  }
  \toprule
    Link & Travel time  \\
         \toprule
     $0\rightarrow1$& 0 \\   
     $1\rightarrow2$& 1\\
     $2\rightarrow3$& 2\\
    $2\rightarrow4$& 1 \\
    $4\rightarrow3$& 1 \\
    $3\rightarrow5$& 1 \\
    $4\rightarrow5$& 2 \\
    $1\rightarrow5$& 4 \\
    $5\rightarrow6$& 0 \\
      \bottomrule
  \end{tabular}               
\end{table}

\begin{table}[htbp]
  \centering
  \caption{Number of observations choosing each path before and after removing link $4\rightarrow5$}
  \label{c4sc1data}
    
  \fontsize{8pt}{10pt}\selectfont
  \begin{tabular}{c c c }
  \toprule
     & Before & After\\
         \toprule
     Path 1&19&25\\ 
     Path 2& 14& 24\\
     Path 3&19&--\\
     Path 4&48&51\\
     
      \bottomrule
  \end{tabular}               
\end{table}

Three models are trained: RL, Res-RL ($M=1$, $\lambda=0$), and ResDGCN-RL ($M=1$, $\lambda=0$). The deterministic utility of RL and systematic component of Res-RL and ResDGCN-RL is defined as $v(i\to j\mid k\to i)=\beta_t\, t_{ij}$ and $\beta_t$ is initialized as $-1$. The residual weight matrices $\theta$ in Res-RL and ResDGCN-RL are initialized as zero matrices. The coefficients $\alpha = \beta = \gamma$ are initialized to 1, and all other parameters are initialized to zero. Model training is conducted using stochastic gradient descent (SGD) with a learning rate of 1e-1 and a maximum of 1000 iterations.

Table~\ref{tab:R3} and Table~\ref{tab:ill_remove} show the estimation results and the predicted choice probabilities (before and after removal) of RL, Res-RL, and ResDGCN-RL. Because RL satisfies the IIA property, after removing link $4\to 5$, the ratios among the remaining paths remain unchanged, implying that these ratios are independent of Path~3. Res-RL, as shown above, relaxes the IIA property and can therefore yield different choice probabilities across paths. However, the choice probability ratio between Path~1 and Path~4 does not depend on link $4\to 5$ because this removal only affects the action from $2\to 4$ to $4\to 3$ through \textit{intersection cross-effect}s. Therefore, the choice probability ratio between Path 1 and Path 4 remains unchanged, regardless of the training data. In contrast, ResDGCN-RL can fit the data because the removal of link $4\to 5$ can also affect the action from $2\to 3$ to $3\to 5$ through \textit{outgoing cross-effect}s, thereby influencing Path~1. If the number of hidden layers of ResDGCN-RL is increased to 2, the removal first affects the action from $1\to 2$ to $2\to 4$ through \textit{neighbor cross-effects}, and then affects the action from $0\to 1$ to $1\to 5$ through \textit{neighbor cross-effect}s, thereby influencing Path~4.

\begin{table}[htbp]
\centering
\caption{Estimation result for illustrative example.}
\label{tab:R3}
\begin{tabular}{lccc}
\toprule
Parameter & RL & Res-RL & ResDGCN-RL \\
\midrule
$\beta_t$        & $-1.000$  & $-0.608$  & $-0.863$ \\
$\alpha$         & --        & --        & $0.994$  \\
$\beta$          & --        & --        & $0.997$  \\
$\gamma$         & --        & --        & $0.996$  \\
\midrule
log-likelihood   & $-248.491$ & $-229.416$ & $-229.112$ \\
\bottomrule
\end{tabular}
\end{table}

\begin{table}[htbp]
\centering
\caption{Choice probabilities before and after removing link $4\rightarrow5$, where the values in parentheses indicate the relative increase. Notice that the change rates of Paths 1 and 4 in Res-RL are identical. This is due to the missing cross-effect capability. On the other hand, ResDGCN-RL accurately predicts all cases.}
\label{tab:ill_remove}
\fontsize{8pt}{10pt}\selectfont
\begin{tabular}{l cc cc cc cc}
\hline
\multirow{3}{*}{Paths}
& \multicolumn{2}{c}{\multirow{2}{*}{Observed choice probability}}

& \multicolumn{6}{c}{Predicted choice probability} \\
\cline{4-9}
&  & 
& \multicolumn{2}{c}{RL}
& \multicolumn{2}{c}{Res-RL}
& \multicolumn{2}{c}{ResDGCN-RL} \\
\cline{2-9}
& Before & After
& Before & After
& Before & After
& Before & After \\
\hline
Path 1 & 19.00\% & \makecell{25.00\%\\(+31.57\%)} 
       & 25.00\% & \makecell{33.33\%\\(+33.33\%)} 
       & 21.18\% & \makecell{23.85\%\\(+12.61\%)} 
       & 19.00\% & \makecell{25.00\%\\(+31.57\%)} \\
Path 2 & 14.00\% & \makecell{24.00\%\\(+71.43\%)} 
       & 25.00\% & \makecell{33.33\%\\(+33.33\%)} 
       & 14.05\% & \makecell{24.60\%\\(+75.09\%)} 
       & 14.00\% & \makecell{24.00\%\\(+71.43\%)}\\
Path 4 & 48.00\% & \makecell{51.00\%\\(+6.25\%)} 
       & 25.00\% & \makecell{33.33\%\\(+33.33\%)} 
       & 45.77\% & \makecell{51.55\%\\(+12.61\%)} 
       & 48.00\% & \makecell{51.00\%\\(+6.25\%)}  \\
\hline
\end{tabular}
\end{table}

\section{Case study}
In this section, the proposed models are validated using a real-world travel dataset from Tokyo. The objectives are fivefold: (i) to verify that the models can be executed and converge stably on a large-scale network, which is challenging for RL because of the required matrix inversion; {\color{red}(ii) to evaluate the benefits of the hybrid models through comparisons with RL, LS-RL, and NRL, and to examine the additional contribution of the GNN through the comparison between Res-RL and ResDGCN-RL;} (iii) to evaluate model generalization under both temporal and destination holdout settings; (iv) to assess interpretability by examining whether the signs and magnitudes of the systematic-component parameters align with behavioral expectations; and (v) to investigate sensitivity to the penalty coefficient $\lambda$ and model depth $M$. {\color{red}As achieving state-of-the-art prediction performance is not the objective of this study, comparisons based solely on prediction accuracy against such models are not considered meaningful for assessing the contribution of the hybrid formulation.}

\subsection{Data description}
\begin{figure}[htbp]
  \centering
  \includegraphics[width=300pt]{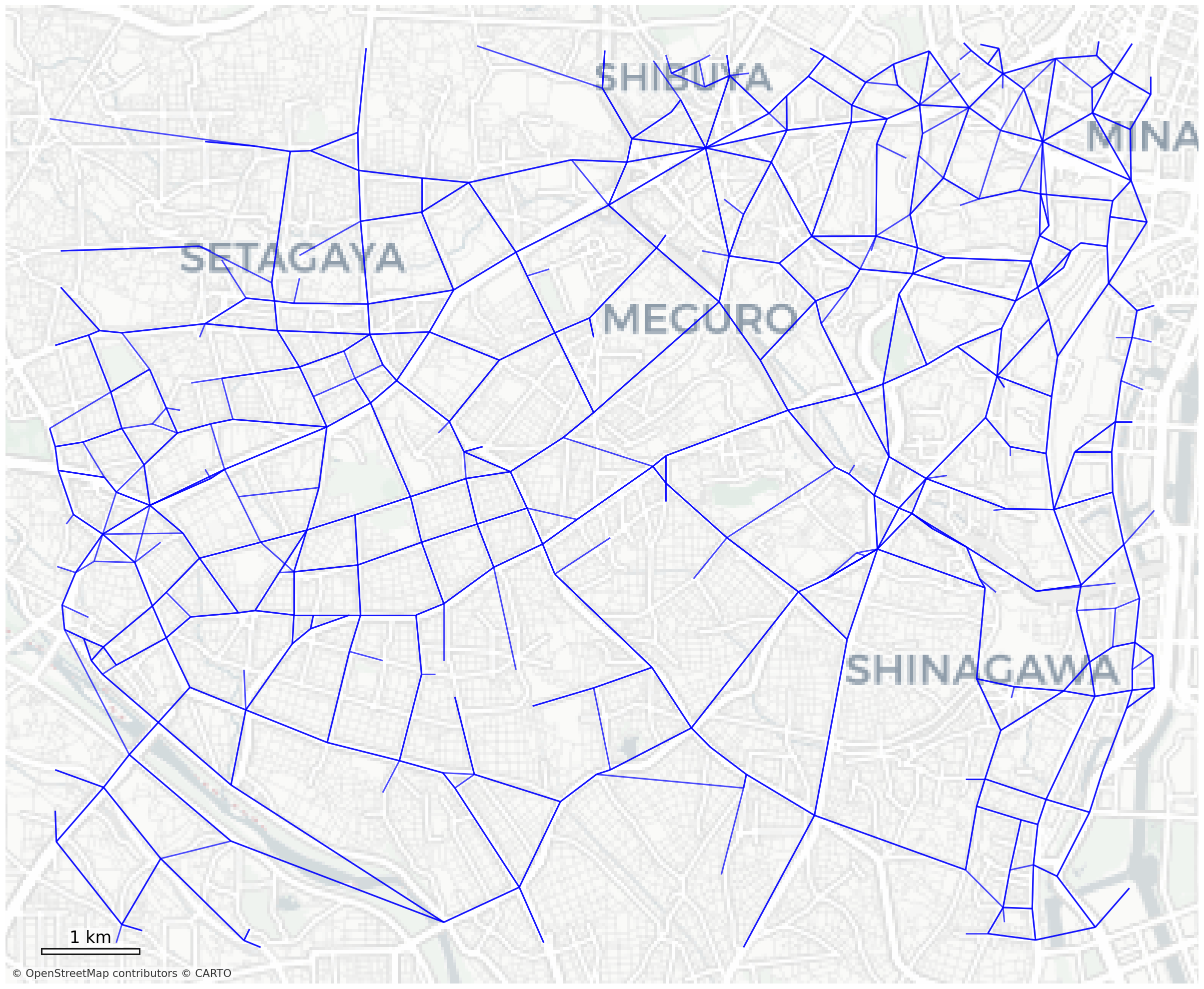}
  \caption{An aggregate road network. Source: © OpenStreetMap contributors, available under the Open Database License (ODbL).}
  \label{network}
\end{figure}

In this section, the dataset used for model validation is introduced. The dataset is based on an aggregated road network consisting of 1,026 links and 375 nodes, together with 547,222 trajectories generated from vehicle trajectory data using the method proposed by \cite{zhonga2023generation}.  The original vehicle trajectory data were provided by the Tokyo Metropolitan Government and were collected in a specific area of Tokyo. As shown in Figure \ref{network}, the resulting network contains only major roads. {A brief description of the road-network aggregation method is provided in Appendix~D. In addition, time-varying link features are used in the experiments.}

\begin{figure}[htbp]
  \centering
  \includegraphics[width=400pt]{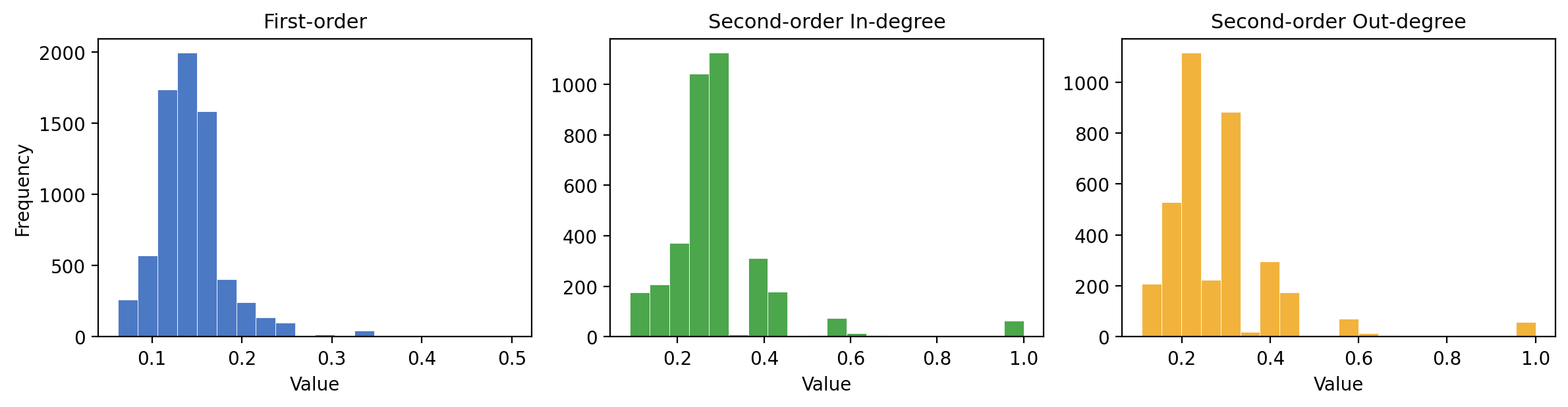}
  \caption{Distribution of nonzero elements in Normalized First-order and Second-order proximity.}
  \label{proxi}
\end{figure}

Figure \ref{proxi} illustrates the distributions of non-zero values in the three normalized proximities. It can be seen that the first-order proximity contains more non-zero values. Moreover, the non-zero values of all three proximities are concentrated around 0.2.

\begin{figure}[htbp]
  \centering
  \includegraphics[width=300pt]{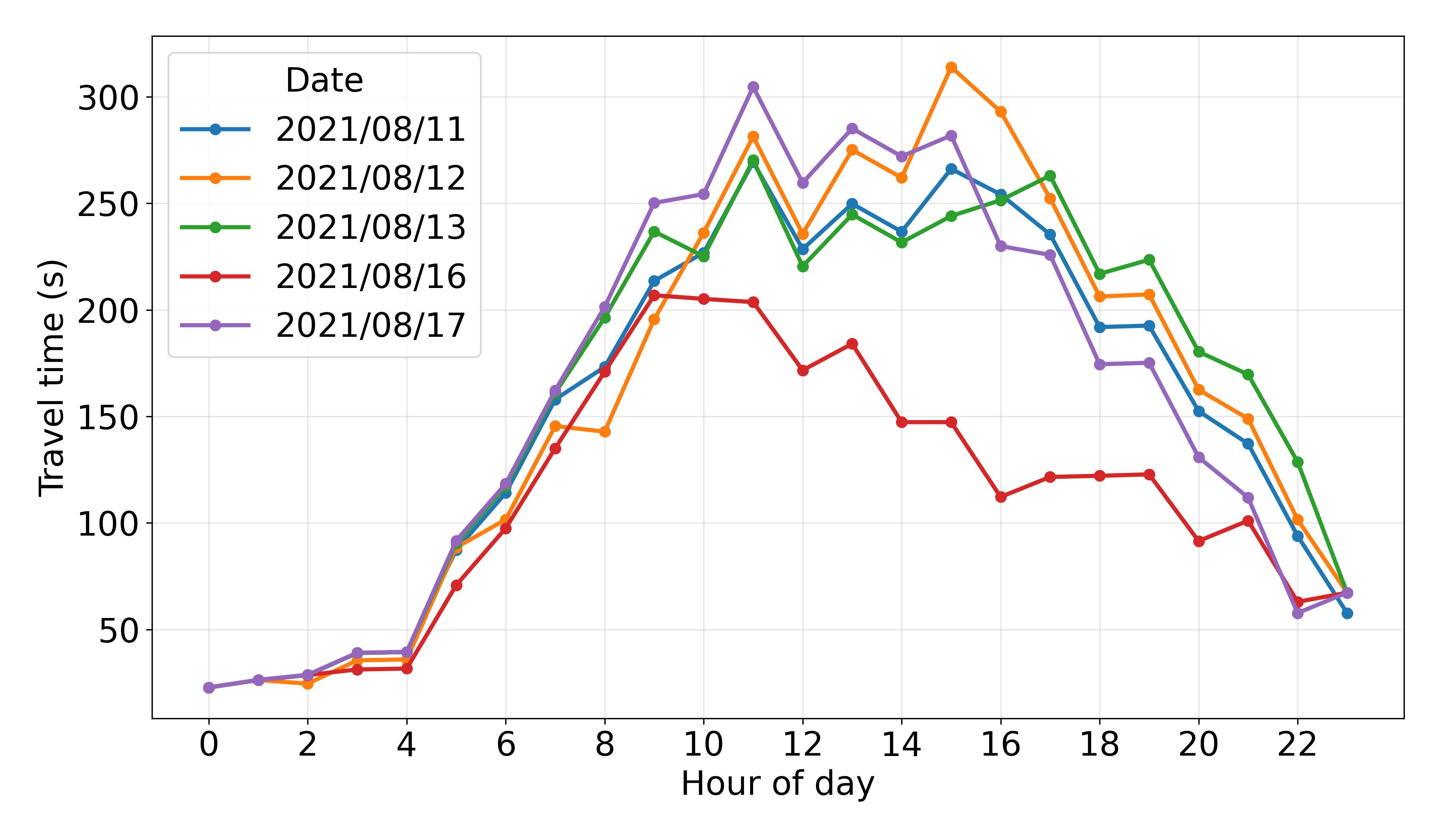}
  \caption{Visualization of time-varying travel time of link 884.}
  \label{tt}
\end{figure}

{To evaluate performance, trajectory data collected over five consecutive weekdays are used. The first three days, comprising 323{,}023 trajectories, are used for training; the fourth day, containing 106{,}953 trajectories, is used for validation; and the fifth day, containing 117{,}246 trajectories, is used for testing. The overlap ratio of unique OD pairs between the training set and each of the validation and test sets is approximately 0.5 for each hourly interval. This indicates that a substantial proportion of OD pairs in the validation and test sets is not observed during training, providing a more challenging evaluation of model generalization.

The use of time-varying link features further enables the evaluation of model generalization. For example, Figure~\ref{tt} shows that the travel time of one selected representative link is lower on August 16, which belongs to the validation set. This dataset split allows us to evaluate whether the model can capture changes in route choice probabilities when the travel time of the link differs from that observed in the training data. Therefore, the cross-day split assesses whether the learned relationships between time-varying traffic conditions and route-choice behavior generalize to different days. } 

\subsection{Model specification and evaluation metrics}
The deterministic utility function for traveling to link $a$ from link $k$ in RL is defined as:
\begin{equation}
  v^{\textrm{RL}}(a|k) = \beta_{\textrm{TT}} \textrm{TT}_a + \beta_{\textrm{Spd}}\textrm{Spd}_a + \beta_{\textrm{RT}} \textrm{RT}_{a|k} + \beta_{\textrm{LC}} + \beta_{\textrm{uturn}}\textrm{uturn}_{a|k}
\end{equation}
where $\mathrm{TT}_a$ and $\mathrm{Spd}_a$ denote the daily average travel time and speed of link $a$, respectively; $\mathrm{RT}_{a\mid k}$ is a right-turn indicator (equal to 1 if the deflection angle from $k$ to $a$ lies between $40^\circ$ and $170^\circ$), and $\mathrm{UTurn}_{a\mid k}$ is a U-turn indicator (equal to 1 if the angle lies between $170^\circ$ and $190^\circ$); $\beta_{\mathrm{LC}}$ is a link-independent constant that penalizes routes with excessive intersections (i.e., a per-link count penalty), while $\beta_{\mathrm{UTurn}}$ is fixed at $-20$ to strongly discourage U-turns; the parameters $\beta_{\mathrm{TT}}$, $\beta_{\mathrm{Spd}}$, $\beta_{\mathrm{RT}}$, and $\beta_{\mathrm{LC}}$ are estimated from the data. The speed feature is included to capture link-level and roads with higher level typically exhibit higher speeds—and, for interpretability and better numerical conditioning, travel time is expressed in minutes and speed in kilometers per minute so that feature magnitudes are on the order of one. The utility functions in Res-RL, and ResDGCN-RL share the same structure as that of RL.

The LS-RL model extends RL by incorporating the link size attribute:
\begin{equation}
  v^{\textrm{LS-RL}}(a|k) = v^{\textrm{RL}}(a|k) + \beta_{\textrm{LS}}\textrm{LS}_{a}
\end{equation}
where $\mathrm{LS}{a}$ is computed using fixed coefficients $\beta{\mathrm{TT}}=-1$, $\beta_{\mathrm{Spd}}=0.5$, $\beta_{\mathrm{RT}}=-0.1$, and $\beta_{\mathrm{LC}}=-0.5$, which are close to the estimates obtained from the RL model.

For NRL, a scale parameter $\mu_k$ is introduced:
\begin{equation}
  \mu_k = \textrm{exp}(\gamma_{\textrm{TT}} \textrm{TT}_a + \gamma_{\textrm{Spd}}\textrm{Spd}_a + \gamma_{\textrm{LS}}\textrm{LS}_{a})
\end{equation}

and the utility functions are given by:
\begin{equation}
    v^{\textrm{NRL}}(a|k) = \frac{1}{\mu_k} v^{\textrm{RL}}(a|k), \quad
\end{equation}

Models are trained with AdamW (learning rate $10^{-5}$, weight decay $10^{-5}$), a batch size of 1{,}024, and up to 200 epochs. Early stopping with a patience of 10 monitored on validation loss is employed to prevent overfitting, and the checkpoint with the best validation performance is reported. Training time depends mainly on the number of trajectories and their length. {For both two-layer Res-RL and ResDGCN-RL, one training epoch takes approximately 4 minutes on an NVIDIA T4 GPU (16 GB), and each additional layer increases the runtime by approximately one minute per epoch.}
 Compared with ResLogit, our RL-based models involve matrix inversion during value-function evaluation and are therefore more numerically delicate to train, which makes initialization particularly critical. Initialization is model-specific. For RL, $\beta_{\mathrm{TT}}=-1.0$, $\beta_{\mathrm{Spd}}=0.0$, $\beta_{\mathrm{RT}}=0.0$, and $\beta_{\mathrm{LC}}=-1.0$. $\beta_{\mathrm{LS}}$ in RL-LS and the parameters in scale parameter are set to zero in NRL. For the hybrid models Res-RL and ResDGCN-RL the systematic component coefficients are initialized with the RL estimates, and the residual component weight matrices are initialized as zero matrices.

{In addition to log-likelihood, three prediction performance metrics—average choice probability (ACP), Jensen–Shannon divergence (JSD), and BLEU—are reported and defined as follows.
ACP measures the mean probability that the model assigns to the observed trajectories and is sensitive to the trajectory length:
\begin{equation}
\label{eq:acp}
\mathrm{ACP}
= \frac{1}{N}\sum_{n=1}^{N}\prod_{l=1}^{\,l_n-1} P^{d}\!\big(a^{n}_{l+1}\mid a^{n}_{l}\big).
\end{equation}
JSD measures the divergence between the empirical next-link distribution and the model-predicted next-link distribution at the predecessor link encountered at each observed step, averaged over the testing set. For a predecessor link $k$, let $\(\hat{p^{d}}(\cdot\mid k)\)$ denote the empirical distribution of next links estimated from validation transitions starting at $k$ {\color{red}in the validation set}, and let $\(p^{d}(\cdot\mid k)\)$ denote the model-predicted distribution at $k$ (conditional on destination $d$. The JSD is defined as:

\begin{equation}
\label{eq:jsd}
\mathrm{JSD}
= \frac{1}{N}\sum_{n=1}^{N}\frac{1}{l_n-1}\sum_{l=1}^{l_n-1} \mathrm{JSD}\!\big(\hat{p^{d}}(\cdot\mid a^{n}_{l})\,\|\,p^{d}(\cdot\mid a^{n}_{l})\big),\quad
\mathrm{JSD}(P\|Q) = \tfrac{1}{2}\mathrm{KL}(P\|M)+\tfrac{1}{2}\mathrm{KL}(Q\|M),
\end{equation}
where $M=\tfrac{1}{2}(P+Q)$ and $\mathrm{KL}(P|Q)=\sum_x P(x)\log\big(P(x)/Q(x)\big)$. Base-2 logarithms are used so that $\mathrm{JSD}\in[0,1]$, where a value of $0$ indicates that the two distributions are identical and larger values reflect greater divergence.

Similarity between the model-predicted most probable route and the observed route is evaluated using sentence-level BLEU-4 with uniform n-gram weights (0.25, 0.25, 0.25, 0.25), following \citet{papineni2002bleu}. BLEU scores range from 0 to 1, with higher values indicating greater similarity and a score of 1 representing identical routes. This metric is useful because the predicted route may differ only slightly from the observed route, and BLEU can still capture their partial similarity.}

\subsection{Estimation result}
\begin{table}[htbp]
\centering
\setlength{\tabcolsep}{3pt} 
\fontsize{8pt}{10pt}\selectfont
\caption{Parameter estimates (standard errors in parentheses) and model performance comparison across models. }

\label{tab:param_estimates}
\begin{tabular}{lccccc}
\toprule
Parameter 
 & RL 
 & RL-LS 
 & NRL 
 & \makecell{Res-RL\\$(M=2,\lambda=0)$} 
 & \makecell{ResDGCN-\\RL $(M=2,\lambda=0)$} 
 \\
\midrule
$\beta_{\mathrm{TT}}$  
 & \makecell{-0.737 \\ (0.002)} 
 & \makecell{ -0.788 \\ (0.002)} 
 & \makecell{-0.762 \\ (0.005)} 
 & \makecell{-0.603 \\ (0.006)} 
 & \makecell{-0.600 \\ (0.005)} 
 \\
$\beta_{\mathrm{Spd}}$ 
 & \makecell{0.087 \\ (0.007)} 
 & \makecell{0.221 \\ (0.005)} 
 & \makecell{0.396 \\ (0.007)} 
 & \makecell{-0.007$^{*}$  \\ ( 0.023)} 
 & \makecell{-0.023$^{*}$ \\ (0.011)} 
 \\
$\beta_{\mathrm{RT}}$  
 & \makecell{-1.610 \\ (0.003)} 
 & \makecell{-1.733 \\ (0.004)} 
 & \makecell{-1.655 \\ (0.010)} 
 & \makecell{-1.605 \\ (0.008)} 
 & \makecell{-1.581 \\ (0.006)} 
 \\
$\beta_{\mathrm{LC}}$  
 & \makecell{-0.139 \\ (0.005)} 
 & \makecell{-0.152 \\ (0.004)} 
 & \makecell{-0.318 \\ (0.006)} 
 & \makecell{-0.136 \\ (0.016)} 
 & \makecell{-0.102 \\ (0.008)} 
\\
$\beta_{\mathrm{LS}}$  
 & -- 
 & \makecell{-0.172 \\ (0.002)} 
 & -- & -- & --  \\
$\gamma_{\mathrm{TT}}$ 
 & -- & -- 
 & \makecell{0.118 \\ (0.001)} 
 & -- & -- \\
$\gamma_{\mathrm{Spd}}$ 
 & -- & -- 
 & \makecell{-0.102 \\ (0.005)} 
 & -- & --  \\
$\gamma_{\mathrm{LS}}$  
 & -- & -- 
 & \makecell{-0.017 \\ (0.001)} 
 & -- & --  \\
\midrule
$\alpha$ 
 & -- & -- & -- & -- & 1.103  \\
$\beta$ 
 & -- & -- & -- & -- & 1.056 \\
$\gamma$ 
 & -- & -- & -- & -- & 1.120 \\
\midrule
Log-likelihood
& -173418.559 & -163370.436 & -150385.923 & -64133.562 & -62726.610 \\
Penalty       
 & 0          & 0          & 0          & 3.885          & 8.874      \\
ACP                    
 & 0.526      & 0.547      & 0.557      & 0.756      & 0.761       \\
JSD                    
 & 0.039      & 0.035      & 0.031      & 0.012      & 0.012       \\
BLEU                   
 & 0.713      & 0.730      & 0.744      & 0.852      & 0.853       \\
Number of parameters   
 & 4          & 5          & 7          & 7158       & 77569      \\
\bottomrule
\end{tabular}

\vspace{2pt}
\begin{minipage}{0.95\linewidth}
\footnotesize $^{*}$\,p$\geq$0.01.
\end{minipage}
\end{table}

Table \ref{tab:param_estimates} reports the estimated parameters for all models, the corresponding standard errors computed from the Fisher information matrix, and the evaluation metrics. It can be observed that both Res-RL and ResDGCN-RL outperform RL and its variants, and ResDGCN-RL further improves upon Res-RL.  ResDGCN-RL introduces substantially more parameters than Res-RL; however, the risk of overfitting is mitigated through early stopping and weight decay. {\color{red}
The number of residual weight parameters is determined by counting the nonzero entries of $\theta^{r,m}$ or $\theta^{d,m}$ across all residual layers. As explained in Section~\ref{resd_property}, the additional parameters in ResDGCN-RL capture cross-effects that Res-RL cannot represent. We therefore interpret the modest improvement over Res-RL as empirical support for the contribution of the GNN module.
} The learned systematic parameters show that they are behaviorally plausible. Most of the systematic component coefficients are statistically significant, which supports the reliability of the behavioral interpretations. {Moreover, across repeated training runs with different random seeds for mini-batch shuffling, the estimated systematic parameters differ by less than 0.001, indicating strong run-to-run stability.} The systematic coefficients exhibit stable signs that are consistent with behavioral expectations. $\beta_{\mathrm{TT}}$ is negative, indicating that travelers prefer routes with shorter travel times. $\beta_{\mathrm{Spd}}$ is positive, showing a preference for higher-level and higher-speed links. $\beta_{\mathrm{RT}}$ is negative, suggesting that travelers tend to avoid right-turn movements, which may involve additional delay and maneuvering complexity due to conflicts with opposing traffic. Finally, $\beta_{\mathrm{LC}}$ is negative, confirming that routes involving more intersections are penalized.

\begin{figure}[htbp]
  \centering
  \includegraphics[width=200pt]{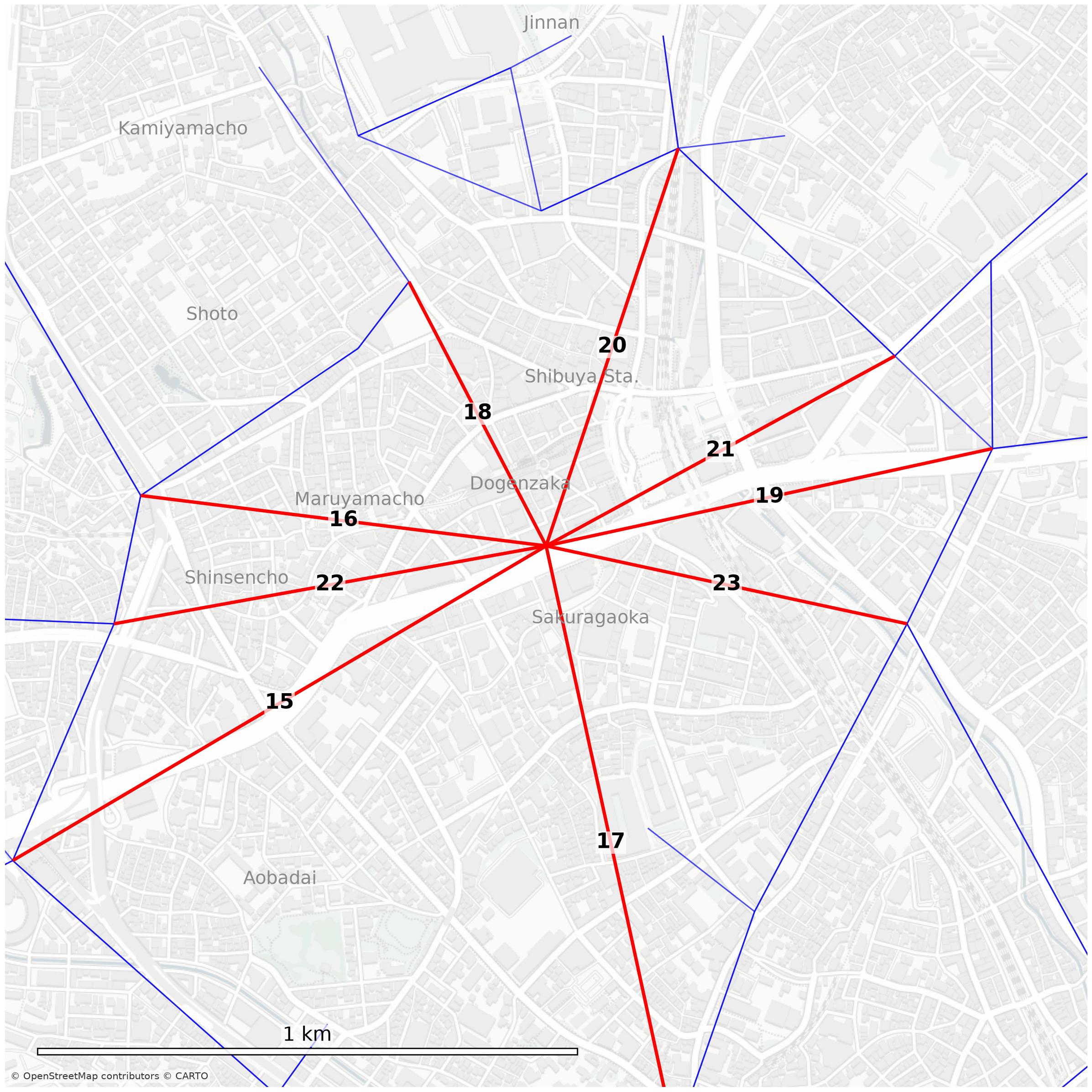}
  \caption{Visualization of the eight links selected for cross-effect analysis. The highlighted red segments represent the target links with their IDs labeled. Note that, in order to enable effective analysis, intersections that are too close to each others are properly aggregated to one node considering their usage patterns in the data using the method of Zhong et al. (2023).
}
  \label{weightmap}
\end{figure}

\begin{figure}[htbp]
  \centering
  \includegraphics[width=400pt]{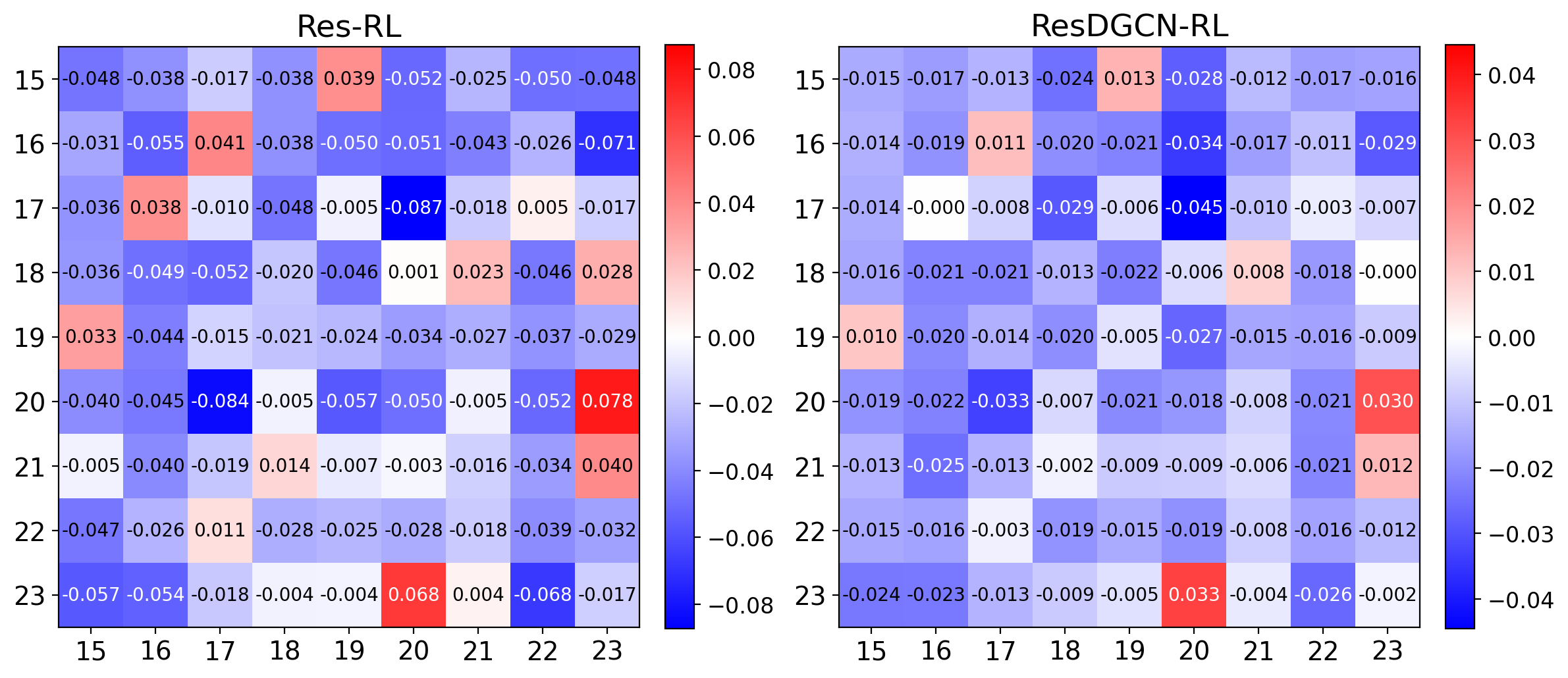}
  \caption{Visualization of $\theta^{r,0}$ and $\theta^{d,0}$ submatrices (rows 15–23, columns 15–23).}
  \label{matrixcompare}
\end{figure}

An intersection is selected (shown in Figure~\ref{weightmap}) that contains eight outgoing links to examine the \textit{intersection cross-effect} learned by Res-RL and ResDGCN-RL. Figure~\ref{matrixcompare} presents the submatrices of $\theta^{r,0}$ and $\theta^{d,0}$ from the 15-th to the 23-rd row and column. A comparison shows that almost all of the corresponding elements share the same sign, suggesting that both models capture largely consistent cross-effect. Nevertheless, a subset of elements exhibit opposite signs, indicating localized discrepancies between the two representations. These differences can be attributed to the fact that $\theta^{d,0}$ in ResDGCN-RL additionally captures other cross-effect. The differences in magnitude can further be explained by the influence of model-specific coefficients such as $\alpha$, $\beta$, and $\gamma$, as well as the normalized proximity in ResDGCN-RL.

Both Res-RL and ResDGCN-RL with $M=2$ were trained under different penalty coefficients. Tables~\ref{tab:param_estimates_rl} and \ref{tab:param_estimates_resdgcnrl} report the estimated parameters of the systematic component, together with the log-likelihood and penalty term of each model. As $\lambda$ increases, prediction accuracy decreases, while the magnitude of the residual component is reduced, reflecting the trade-off between accuracy and interpretability. The sensitivity to $\lambda$ differs between Res-RL and ResDGCN-RL because of their different model structures. When $\lambda=0$, the estimated $\beta_{\mathrm{Spd}}$ is negative in both hybrid models, whereas it is positive in the standard RL model. As $\lambda$ increases to 2, $\beta_{\mathrm{Spd}}$ becomes positive and more consistent with the RL estimate, indicating improved behavioral interpretability. Similarly, the estimated $\beta_{\mathrm{TT}}$ becomes closer to that of RL as the residual component is more strongly penalized. Since $\beta_{\mathrm{TT}}$ is directly related to the valuation of travel time, this increased consistency is particularly important for the behavioral interpretation of the model.

\begin{table}[htbp]
\centering
\setlength{\tabcolsep}{3pt}
\fontsize{7pt}{9pt}\selectfont
\caption{Systematic component parameter estimates (standard errors in parentheses) and performance comparison across Res-RL models with different penalty coefficients.}
\label{tab:param_estimates_rl}
\begin{tabular}{lcccccc}
\toprule
Parameter
 & \makecell{Res\text{-}RL\\$(M=2,\lambda=0)$}
 & \makecell{Res\text{-}RL\\$(M=2,\lambda=0.4)$}
 & \makecell{Res\text{-}RL\\$(M=2,\lambda=0.8)$}
 & \makecell{Res\text{-}RL\\$(M=2,\lambda=1.2)$}
 & \makecell{Res\text{-}RL\\$(M=2,\lambda=1.6)$}
 & \makecell{Res\text{-}RL\\$(M=2,\lambda=2)$} \\
\midrule
$\beta_{\mathrm{TT}}$
 & \makecell{-0.603 \\ (0.006)}
 & \makecell{-0.674 \\ (0.003)}
 & \makecell{-0.693 \\ (0.011)}
 & \makecell{-0.710 \\ (0.008)}
 & \makecell{-0.724 \\ (0.002)}
 & \makecell{-0.736 \\ (0.006)} \\
$\beta_{\mathrm{Spd}}$
 & \makecell{-0.007$^{*}$  \\ (0.023)}
 & \makecell{-0.045 \\ (0.007)}
 & \makecell{-0.035 \\ (0.002)}
 & \makecell{-0.015 \\ (0.003)}
 & \makecell{0.018 \\ (0.004)}
 & \makecell{0.060 \\ (0.006)} \\
$\beta_{\mathrm{RT}}$
 & \makecell{-1.605 \\ (0.008)}
 & \makecell{-1.700 \\ (0.008)}
 & \makecell{-0.167 \\ (0.024)}
 & \makecell{-0.188 \\ (0.013)}
 & \makecell{-1.614 \\ (0.003)}
 & \makecell{-1.619 \\ (0.003)} \\
$\beta_{\mathrm{LC}}$
 & \makecell{-0.136 \\ (0.016)}
 & \makecell{-0.240 \\ (0.008)}
 & \makecell{-0.215 \\ (0.011)}
 & \makecell{-0.187 \\ (0.009)}
 & \makecell{-0.160 \\ (0.007)}
 & \makecell{-0.138 \\ (0.005)} \\
 \midrule
Log-likelihood
& -64133.562 & -71502.867 & -88718.571 & -103081.914 & -116116.002 & -119725.300 \\
Penalty      
 & 3.885         & 0.763          & 0.408          & 0.234          & 0.124     & 0.041 \\
 Number of parameters & \multicolumn{6}{c}{7158} \\
\bottomrule
\end{tabular}

\vspace{2pt}
\begin{minipage}{0.95\linewidth}
\footnotesize  $^{*}$\,p$\geq$0.01.
\end{minipage}
\end{table}

\begin{table}[htbp]
\centering
\setlength{\tabcolsep}{3pt}
\fontsize{7pt}{9pt}\selectfont
\caption{Systematic component parameter estimates (standard errors in parentheses) and performance comparison across ResDGCN-RL models with different penalty coefficients.}
\label{tab:param_estimates_resdgcnrl}
\begin{tabular}{lcccccc}
\toprule
Parameter
 & \makecell{ResDGCN\text{-}RL\\$(M=2,\lambda=0)$}
 & \makecell{ResDGCN\text{-}RL\\$(M=2,\lambda=0.4)$}
 & \makecell{ResDGCN\text{-}RL\\$(M=2,\lambda=0.8)$}
 & \makecell{ResDGCN\text{-}RL\\$(M=2,\lambda=1.2)$}
 & \makecell{ResDGCN\text{-}RL\\$(M=2,\lambda=1.6)$}
 & \makecell{ResDGCN\text{-}RL\\$(M=2,\lambda=2)$} \\
\midrule
$\beta_{\mathrm{TT}}$
 & \makecell{-0.600 \\ (0.005)}
 & \makecell{-0.699\\ (0.002)}
 & \makecell{-0.719 \\ (0.001)}
 & \makecell{-0.731 \\ (0.002)}
 & \makecell{-0.737 \\ (0.002)}
 & \makecell{-0.740 \\ (0.002)} \\
$\beta_{\mathrm{Spd}}$
 & \makecell{-0.023$^{*}$ \\ (0.011)}
 & \makecell{-0.035 \\ (0.007)}
 & \makecell{-0.019$^{*}$ \\ (0.022)}
 & \makecell{0.004$^{*}$ \\ (0.004)}
 & \makecell{0.013$^{*}$ \\ (0.006)}
 & \makecell{0.032 \\ (0.006)} \\
$\beta_{\mathrm{RT}}$
 & \makecell{-1.581 \\ (0.006)}
 & \makecell{-1.708 \\ (0.004)}
 & \makecell{-1.667 \\ (0.004)}
 & \makecell{-1.626 \\ (0.003)}
 & \makecell{-1.596  \\ (0.003)}
 & \makecell{-1.595 \\ (0.005)} \\
$\beta_{\mathrm{LC}}$
 & \makecell{-0.102 \\ (0.008)}
 & \makecell{ -0.244 \\ (0.015)}
 & \makecell{-0.222 \\ (0.018)}
 & \makecell{-0.205\\ (0.003)}
 & \makecell{-0.186 \\ (0.004)}
 & \makecell{-0.165 \\ (0.005)} \\
 \midrule
Log-likelihood         
 & -62726.610 & -68936.664 & -81324.587 & -93140.191 & -103654.232 & -113392.875 \\
Penalty
 & 8.874          & 0.663          & 0.407          & 0.263          & 0.172     & 0.108 \\
Number of parameters & \multicolumn{6}{c}{77569} \\
\bottomrule
\end{tabular}

\vspace{2pt}
\begin{minipage}{0.95\linewidth}
\footnotesize  $^{*}$\,p$\geq$0.01.
\end{minipage}
\end{table}

\begin{figure}[htbp]
  \centering
  \includegraphics[width=200pt]{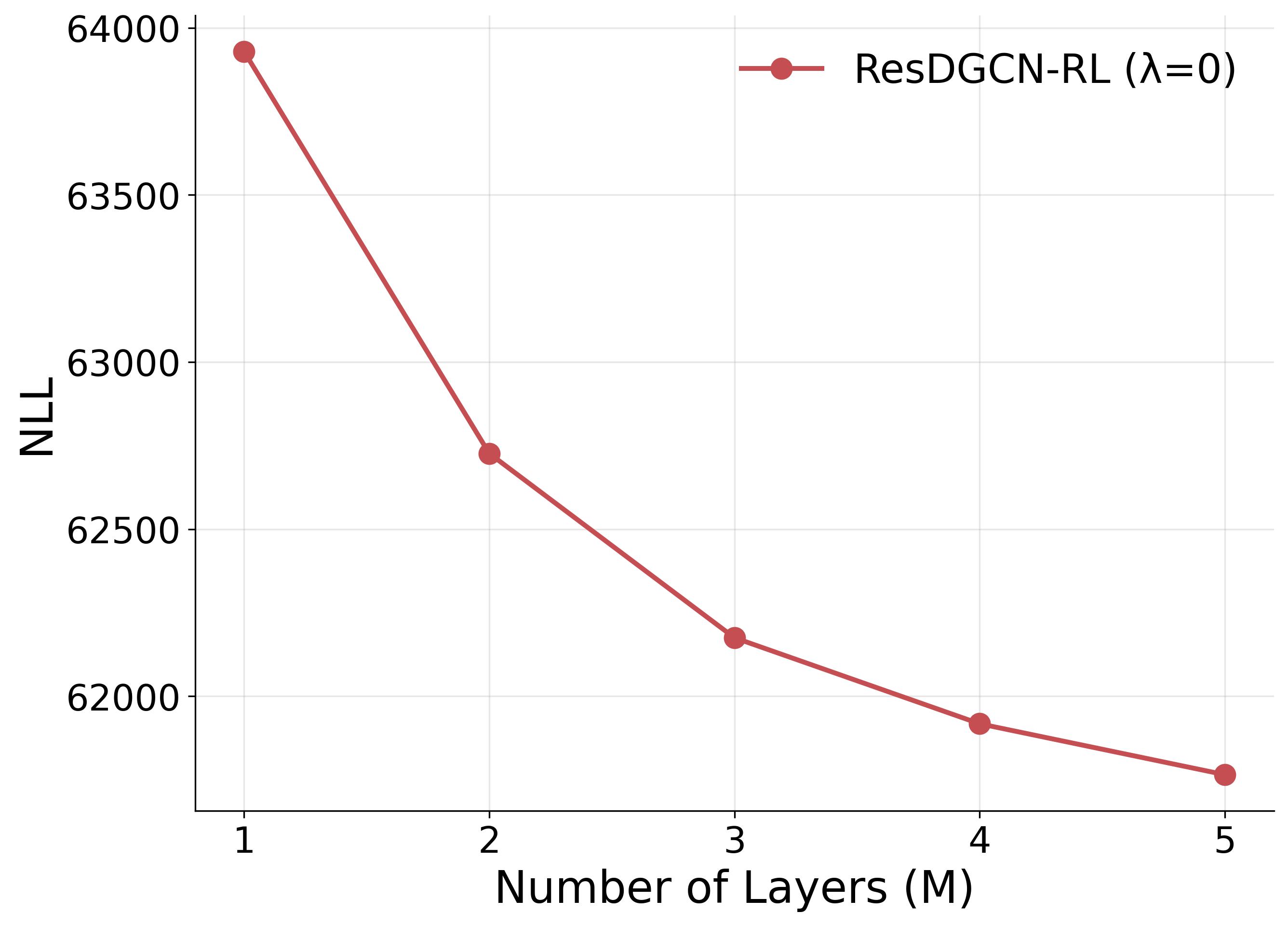}
  \caption{Sensitivity analysis on model depth.}
  \label{fig:nll_depth}
\end{figure}

{
ResDGCN-RL was trained under the setting of $\lambda=0$ with varying model depths. Figure~\ref{fig:nll_depth} presents the NLL values obtained at different depths. The NLL decreases as $M$ increases, indicating that greater model depth improves predictive performance by increasing the expressive capacity of the residual structure. However, the improvement becomes relatively limited as additional layers are introduced.

}

\begin{figure}[htbp]
  \centering

  \begin{subfigure}[b]{0.48\linewidth}
    \centering
    \includegraphics[width=\linewidth,height=4.2cm,keepaspectratio]{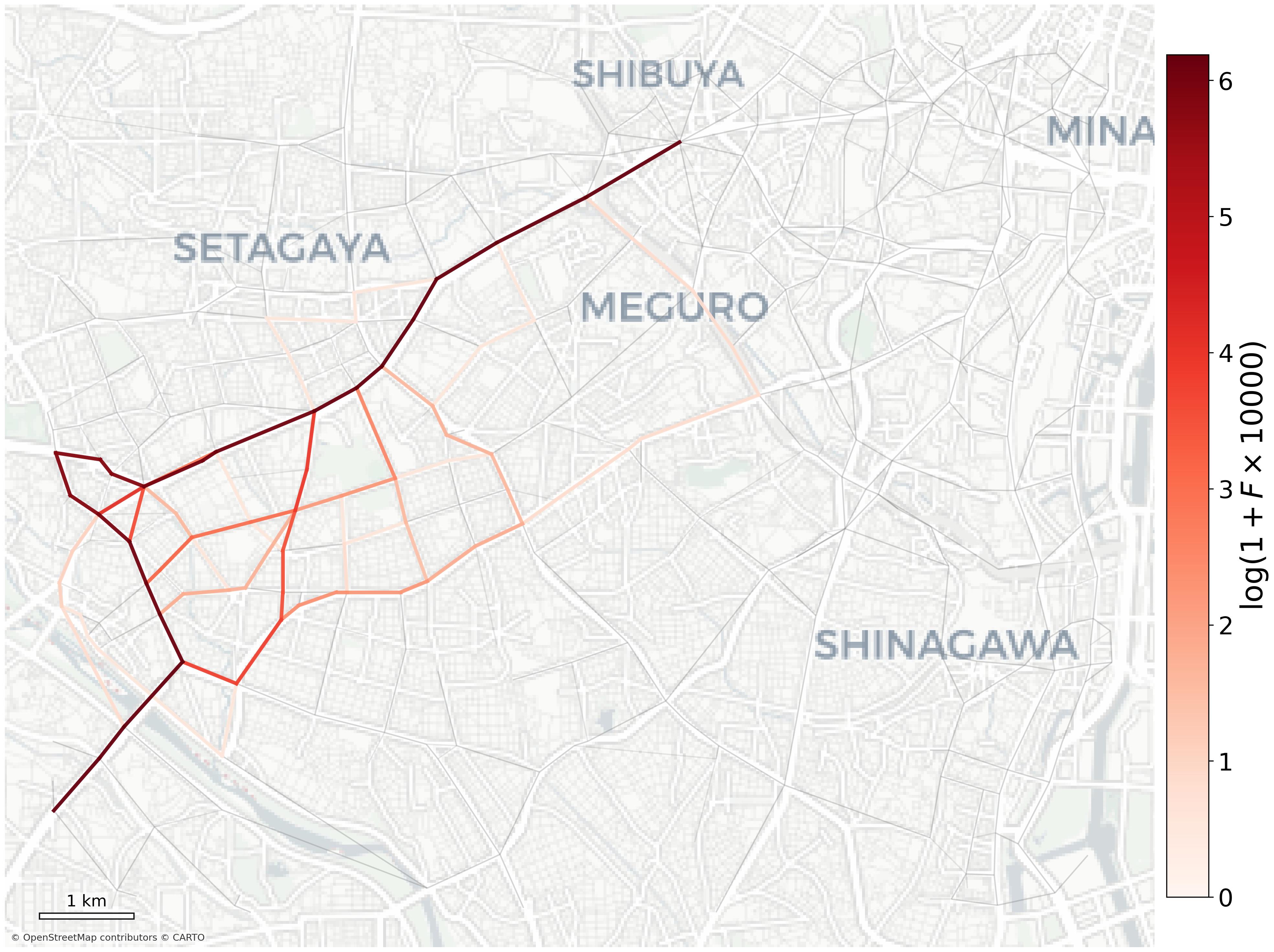}
    \caption{Real distribution}
    \label{fig:cf2-a}
  \end{subfigure}
  \hfill
  \begin{subfigure}[b]{0.48\linewidth}
    \centering
    \includegraphics[width=\linewidth,height=4.2cm,keepaspectratio]{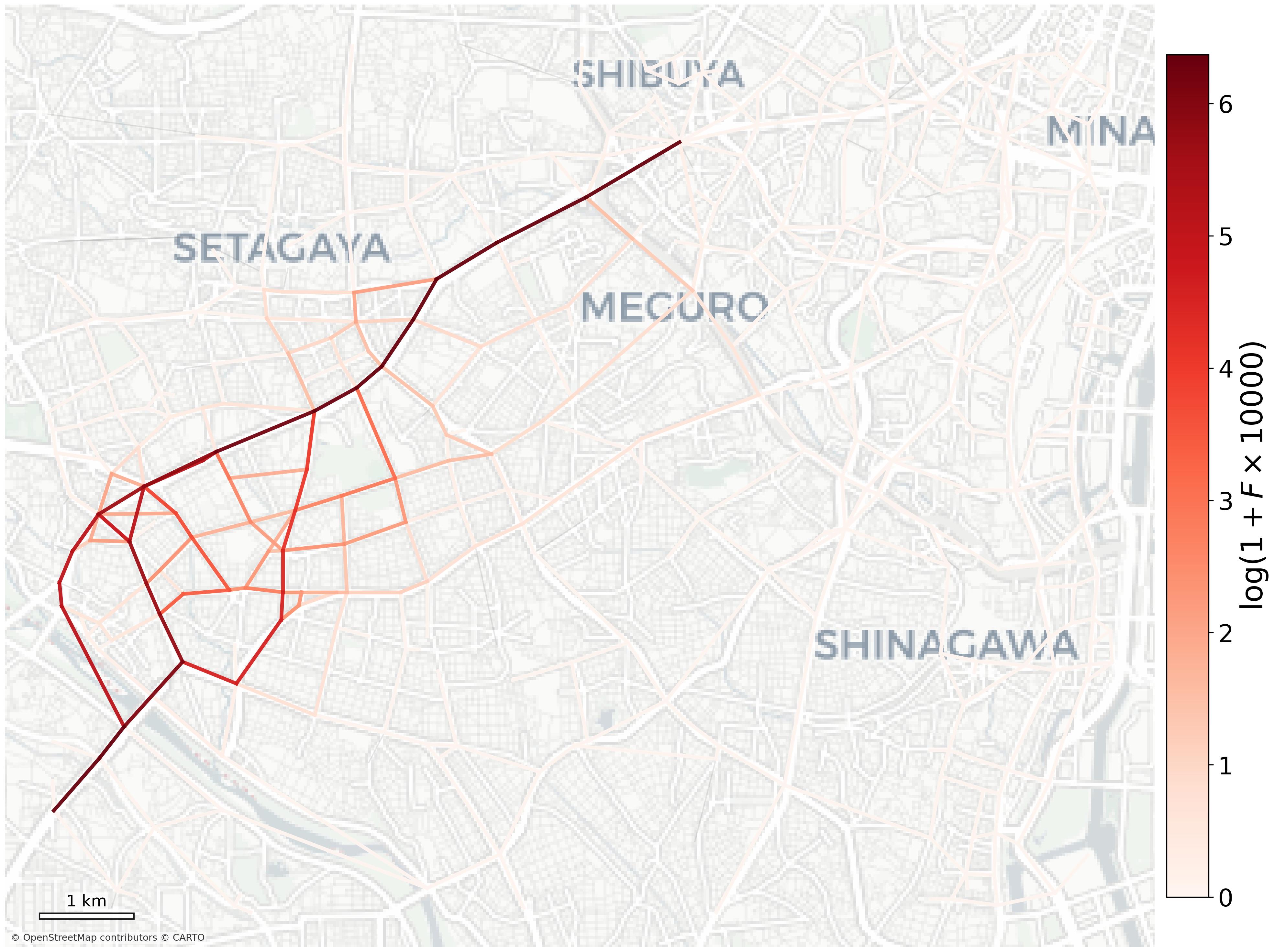}
    \caption{RL}
    \label{fig:cf2-b}
  \end{subfigure}

  \vspace{0.4em}

  \begin{subfigure}[b]{0.48\linewidth}
    \centering
    \includegraphics[width=\linewidth,height=4.2cm,keepaspectratio]{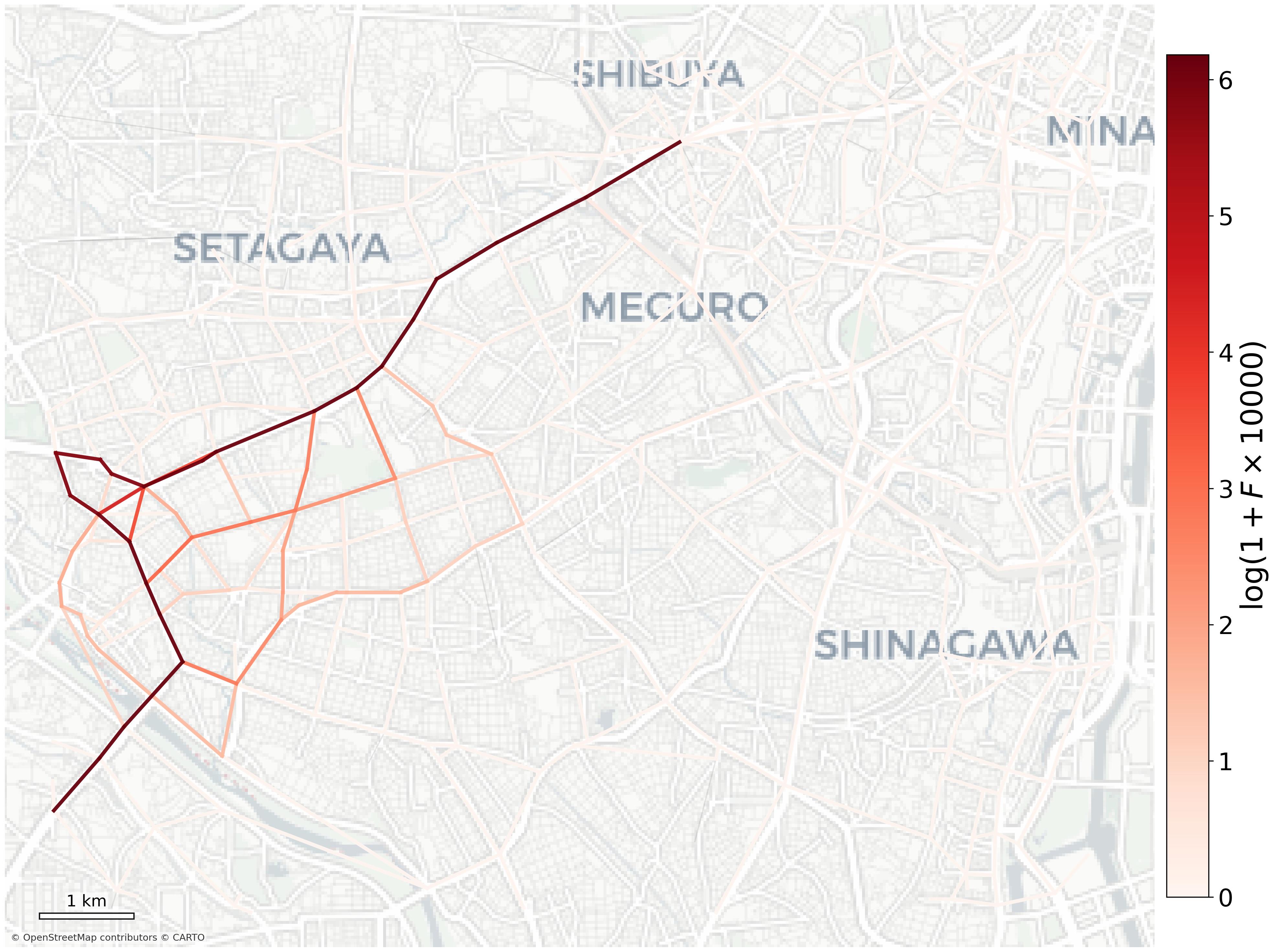}
    \caption{Res-RL ($M=2,\lambda=0$)}
    \label{fig:cf2-c}
  \end{subfigure}
  \hfill
  \begin{subfigure}[b]{0.48\linewidth}
    \centering
    \includegraphics[width=\linewidth,height=4.2cm,keepaspectratio]{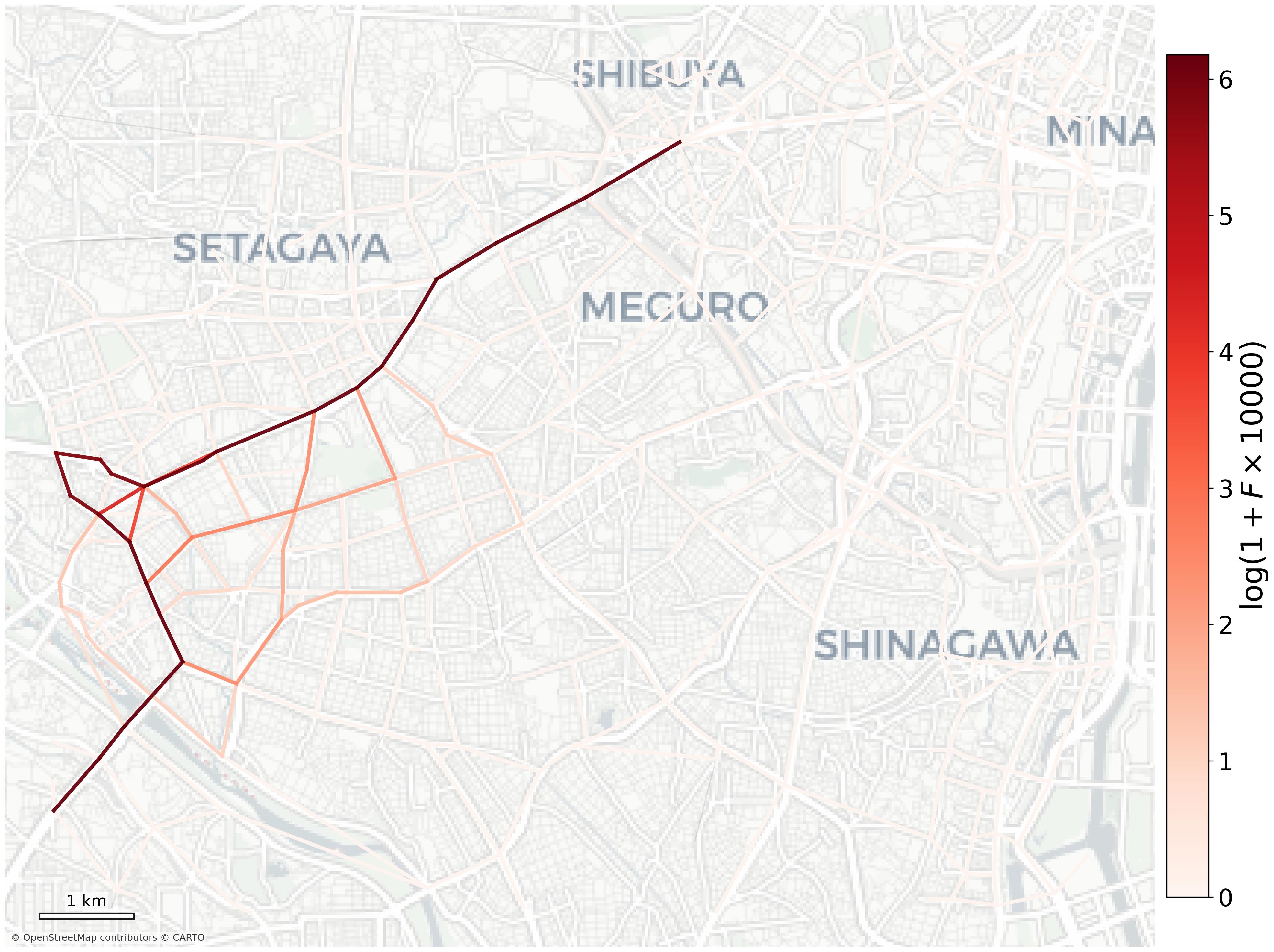}
    \caption{ResDGCN-RL ($M=2,\lambda=0$)}
    \label{fig:cf2-d}
  \end{subfigure}

  \caption{Expected link flow distributions under different models from node 0 to 52 using the link features at 6 p.m. on August 17. The colorbar shows $\log(1+10000\times F)$, where $F$ denotes the expected link flow. }
  \label{fig:linkflow}
\end{figure}

\begin{figure}[htbp]
    \centering
    \includegraphics[width=0.5\linewidth]{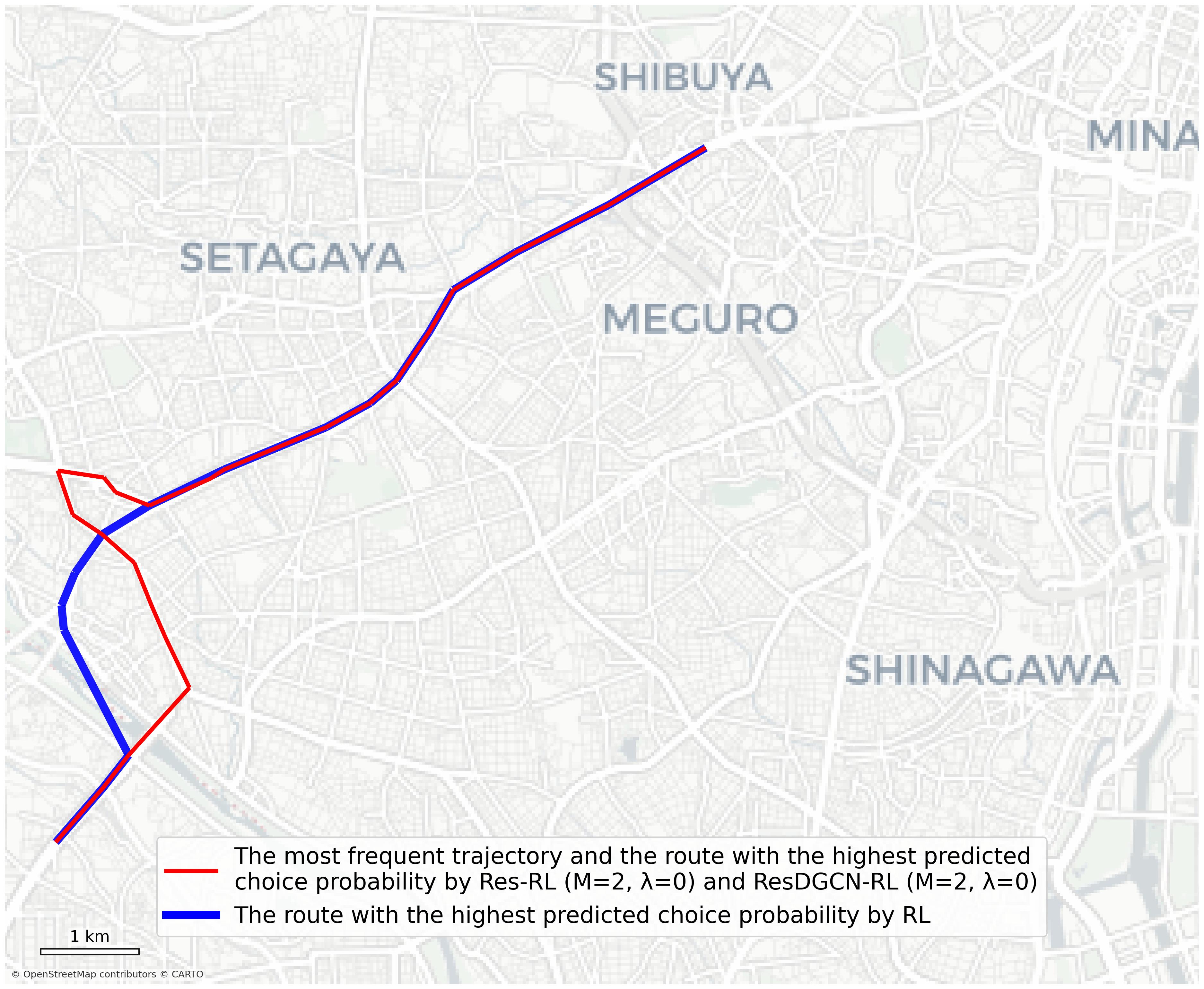}
    \caption{The most frequent trajectory from node 0 to 52 and the route with the highest choice probability predicted by different models.}
    \label{fig:real}
\end{figure}

{
As an illustrative example, one OD pair is selected to compute the expected link flow, with its demand assumed to be one. The expected flow on each link represents the expected usage of that link under the route-choice probabilities predicted by the model. The expected link flows are computed for RL, Res-RL (M=2, $\lambda=0$), and ResDGCN-RL (M=2, $\lambda=0$) using Equation~\eqref{eq:linkflow} and the link features at 6 p.m. on August 17. The resulting distributions are visualized in Figure~\ref{fig:linkflow} after applying a logarithmic transformation. In this example, the distributions predicted by Res-RL and ResDGCN-RL are visually closer to the observed distribution than that predicted by RL. Both models also assign the highest expected flow to the same link as in the observed data.

Figure~\ref{fig:real} compares the most frequent observed trajectory with the routes assigned the highest choice probabilities by the different models for the selected OD pair. The route predicted by the proposed model, shown in red, coincides with the most frequent observed trajectory, whereas the route predicted by RL, shown in blue, differs slightly. Table~\ref{tab:traj2} reports the utilities of these two routes. The residual utility assigned to the red route by the proposed model is higher than that assigned to the blue route, thereby increasing the choice probability of the red route.

\begin{table}[htbp]
\centering
\fontsize{8pt}{10pt}\selectfont
\caption{Systematic and total utility for two routes.}
\label{tab:traj2}
\begin{tabular}{llccc}
\toprule
Route & Component & RL & \makecell{Res-RL \\ $(M=2,\lambda=0)$}  & \makecell{ResDGCN-RL \\ $(M=2,\lambda=0)$} \\
\midrule
\multirow{2}{*}{Route predicted by RL}  & Systematic utility &-21.131  & -18.577 & -17.727 \\
                      & Total instantaneous utility      & -21.131 & -22.562 & -22.812 \\
\midrule
\multirow{2}{*}{Most frequent trajectory} & Systematic utility & -29.464 & -26.006 & -24.836 \\
                      & Total instantaneous utility      &  -29.464 & -18.010 & -17.606 \\
\bottomrule
\end{tabular}
\end{table}

\subsection{Temporal and destination holdout}
Even when the training and test sets are separated by date, substantial similarity in link features and considerable overlap in observed trajectories may remain. This is common in route choice datasets and may make it difficult to assess whether a model can generalize beyond the travel patterns already represented in the training data. Therefore, in addition to the original date-based split, two alternative holdout schemes were constructed, consisting of a temporal holdout and a destination holdout.

For the temporal holdout, trajectories with departure times before 12:00 are assigned to the training set, while trajectories departing at or after 12:00 are randomly divided into the validation and test sets. The resulting training, validation, and test sets contain 249,385, 148,918, and 148,919 trajectories, respectively. This split is designed to evaluate the temporal generalization ability of the models under potential variations in link features across different periods of the day, such as differences between free-flow and congested traffic conditions.

For the destination holdout, the data are divided into training, validation, and test sets with an approximate ratio of 7:2:1, while ensuring that their destination sets are mutually exclusive. The resulting sets contain 385,662, 107,538, and 54,022 trajectories, covering 705, 132, and 120 destinations, respectively. This split is used to evaluate generalization to unseen destinations, which is particularly important for RL-based models because the value function is destination-dependent.

\begin{figure}[htbp]
    \centering
    \includegraphics[width=1\linewidth]{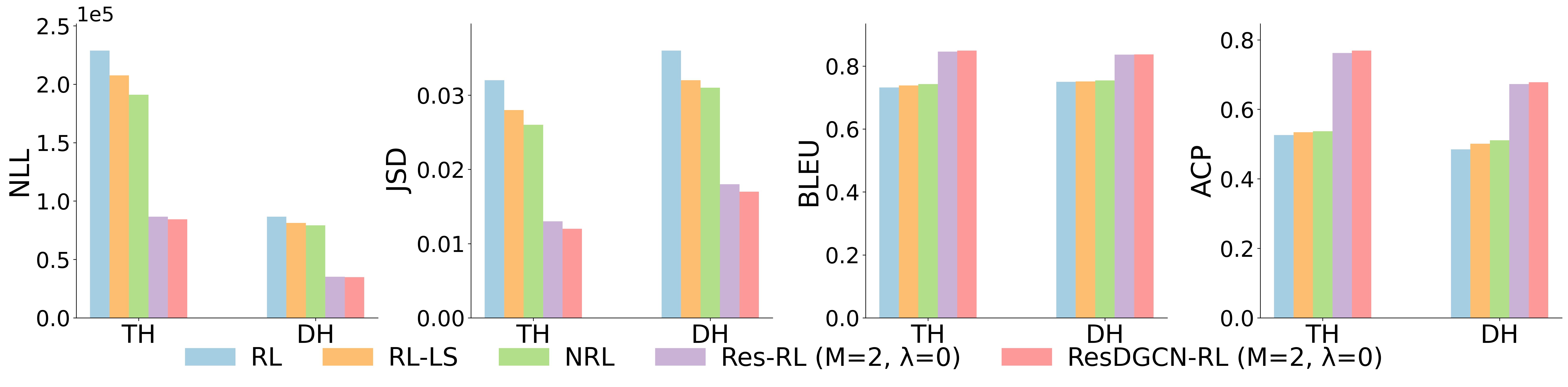}
    \caption{Model performance of temporal holdout (TH) and destination holdout (DH).}
    \label{fig:holdout}
\end{figure}

Figure~\ref{fig:holdout} shows that both Res-RL and ResDGCN-RL consistently outperform the conventional RL-based models across all evaluation metrics under the different holdout settings, confirming the effectiveness of the proposed models. Performance under the destination holdout is generally lower than under the other two data-splitting settings, indicating that prediction for entirely unseen destinations is more challenging. ResDGCN-RL achieves only modest improvements over Res-RL. Nevertheless, as discussed previously, ResDGCN-RL is theoretically more expressive because its graph-based message-passing structure can capture broader substitution patterns beyond the local cross-effects modeled by Res-RL. Detailed parameter estimates are provided in the Appendix~E.}
\section{Conclusion and discussion}
In this study, the ResLogit approach is extended to the RL framework, resulting in a link-based hybrid model termed Res-RL. To better capture the substitution pattern, ResDGCN-RL is further proposed, which integrates RL with DGCN to learn more complex cross-effect patterns. {These models extend RL by decomposing the deterministic utility term into a systematic component specified using a predefined utility function to provide behavioral interpretability and a residual component learned through deep learning techniques to improve predictive performance.
} Moreover, given the importance of interpretability in route choice modeling, a regularization-based loss function is introduced to balance the trade-off between interpretability and prediction accuracy.

Both Res-RL and ResDGCN-RL are proved that they can relax the IIA property. However, Res-RL can only partially capture cross-effects between paths, leading to unrealistic substitution pattern under particular conditions. By message-passing mechansim of GNNs, ResDGCN-RL can fully capture cross-effects between more path pairs.

To evaluate the effectiveness of the proposed models, experiments are conducted on a real-world trajectory dataset from Tokyo. The results show that the proposed models outperform RL and its variants in prediction accuracy, while the estimated systematic parameters remain behaviorally plausible. The effect of the penalty coefficient $\lambda$ on the trade-off between prediction performance and the penalty term is also examined. As additional analyses, temporal holdout and destination holdout experiments are conducted to assess model generalization.

Despite these improvements, this study has several limitations related to both model design and empirical evaluation. {First, when the network is large, a GCN typically requires many hidden layers to propagate information across distant links, making it difficult to capture substitution patterns among paths that are far apart in the graph and potentially leading to unstable results. Experiments on a denser road network are therefore necessary. Second, the space complexity remains high due to the large weight matrix, posing challenges for large-scale applications. {Third, the intended application of the current models is route choice prediction within an existing observed network whose links are already included in the training data. The models can accommodate link removals by updating the adjacency matrix to disable the corresponding transitions, i.e., by setting the $i$-th row and column to zero. However, they cannot directly handle newly added links because no representations or cross-effects have been learned for these links.} Finally, while the systematic component retains an economic interpretation, the residual cross-effects are primarily data-driven and do not necessarily map to a specific behavioral mechanism, unlike in ResLogit-style formulations.

Future research could address these limitations by integrating attention mechanisms \citep{vaswani2017attention} to capture long-range dependencies in road networks, {and exploring inductive architectures, such as GraphSAGE \citep{hamilton2017inductive} to improve model generalization ability across different network structures while reducing the computational cost.}
\section*{Acknowledgment}
The authors would like to thank the editors and reviewers for their kind handling of our manuscript and their insightful and constructive comments that have significantly improved this work. The data is provided by Tokyo Metropolitan Government through their project 'offering of several datasets on mobility and transportation in Tokyo 2020 Games'. Map data copyrighted OpenStreetMap contributors and available from https://www.openstreetmap.org. This work was partly funded by JSPS KAKENHI Grant-in-Aid for Scientific Research 20H02267 and JST/JICA SATREPS JPMJSA2405. 
\section*{CRediT author contribution statement}
\textbf{Ma Yuxun:} Conceptualization; Methodology; Data curation; Formal analysis; Investigation; Validation; Visualization; Writing - original draft. \textbf{Toru Seo:} Conceptualization; Data curation; Funding acquisition; Project administration; Resources; Supervision; Writing - review \& editing.
\section*{Declaration of competing interest}
The authors declare that there are no known competing interests.
\section*{Declaration of generative AI and AI-assisted technologies in the writing process}
During the preparation of this work, the authors used Large Language Models for a grammatical proofreading purpose. After using this tool/service, the authors reviewed and edited the content as needed and take full responsibility for the content of the publication.

\bibliography{route, pt, MNP,rl,clogit,nrl,tf1,tf2,tf3,tf4,ts1,ts2,ts3,sc1,sc2,sc3,st1,st2,av1,av2,choice1,choice2,repr,tst,attc,reslogit,gnn1,gnn2,gnn3,gnn4,gnn5,gnn6,gnn7,gumble,pscl,pcl,mix,gn1,mrl,ymmt,LRT,rf,sch,NN1,NN2,NN3,NN4,rl1,rl2,rnn1,rnn2,rbm,resnet,gcn,dgcn,data,exp1,lrp,att, psl2,PCL_chu,mother_logit,CNL_vov,MNP2,RMM-RL,bleu,perturbed,choicediscuss,setgen,motherlogit,drrl,meta-irl,gnndcm1,gnndcm2,graphsage}

@article{vovsha1997application,
  title={Application of cross-nested logit model to mode choice in Tel Aviv, Israel, metropolitan area},
  author={Vovsha, Peter},
  journal={Transportation Research Record},
  volume={1607},
  number={1},
  pages={6--15},
  year={1997},
  publisher={SAGE Publications Sage CA: Los Angeles, CA}
}

@article{lee2010hybrid,
  title={A hybrid tree approach to modeling alternate route choice behavior with online information},
  author={Lee, Chanyoung and Ran, Bin and Yang, Fan and Loh, Wei-Yin},
  journal={Journal of Intelligent Transportation Systems},
  volume={14},
  number={4},
  pages={209--219},
  year={2010},
  publisher={Taylor \& Francis}
}

@article{daganzo1977stochastic,
  title={On stochastic models of traffic assignment},
  author={Daganzo, Carlos F and Sheffi, Yosef},
  journal={Transportation science},
  volume={11},
  number={3},
  pages={253--274},
  year={1977},
  publisher={INFORMS}
}

@article{yang1993exploration,
  title={Exploration of route choice behavior with advanced traveler information using neural network concepts},
  author={Yang, Hai and Kitamura, Ryuichi and Jovanis, Paul P and Vaughn, Kenneth M and Abdel-Aty, Mohamed A},
  journal={Transportation},
  volume={20},
  pages={199--223},
  year={1993},
  publisher={Springer}
}

@article{dia2007modelling,
  title={Modelling drivers' compliance and route choice behaviour in response to travel information},
  author={Dia, Hussein and Panwai, Sakda},
  journal={Nonlinear Dynamics},
  volume={49},
  pages={493--509},
  year={2007},
  publisher={Springer}
}

@article{politis2023route,
  title={A Route Choice Model for the Investigation of Drivers’ Willingness to Choose a Flyover Motorway in Greece},
  author={Politis, Ioannis and Georgiadis, Georgios and Kopsacheilis, Aristomenis and Nikolaidou, Anastasia and Sfyri, Chrysanthi and Basbas, Socrates},
  journal={Sustainability},
  volume={15},
  number={5},
  pages={4614},
  year={2023},
  publisher={MDPI}
}

@article{lai2019understanding,
  title={Understanding drivers' route choice behaviours in the urban network with machine learning models},
  author={Lai, Xinjun and Fu, Hui and Li, Jun and Sha, Zhiren},
  journal={IET Intelligent Transport Systems},
  volume={13},
  number={3},
  pages={427--434},
  year={2019},
  publisher={Wiley Online Library}
}

@inproceedings{chu1989paired,
  title={A paired combinatorial logit model for travel demand analysis},
  author={Chu, Chaushie},
  booktitle={Proceedings of the fifth world conference on transportation research},
  volume={4},
  pages={295--309},
  year={1989},
  organization={Ventura CA}
}

@article{mai2017similarities,
  title={On the similarities between random regret minimization and mother logit: The case of recursive route choice models},
  author={Mai, Tien and Bastin, Fabian and Frejinger, Emma},
  journal={Journal of choice modelling},
  volume={23},
  pages={21--33},
  year={2017},
  publisher={Elsevier}
}

@article{vaswani2017attention,
  title={Attention is all you need},
  author={Vaswani, Ashish and Shazeer, Noam and Parmar, Niki and Uszkoreit, Jakob and Jones, Llion and Gomez, Aidan N and Kaiser, {\L}ukasz and Polosukhin, Illia},
  journal={Advances in neural information processing systems},
  volume={30},
  year={2017}
}

@article{phan2022attentionchoice,
  title={Attentionchoice: Discrete choice modelling supported by a deep learning attention mechanism},
  author={Phan, Danh T and Vu, Hai L and Currie, Graham},
  journal={Available at SSRN 4305637},
  year={2022}
}

@inproceedings{papineni2002bleu,
  title={Bleu: a method for automatic evaluation of machine translation},
  author={Papineni, Kishore and Roukos, Salim and Ward, Todd and Zhu, Wei-Jing},
  booktitle={Proceedings of the 40th annual meeting of the Association for Computational Linguistics},
  pages={311--318},
  year={2002}
}

@article{van2022choice,
  title={Choice modelling in the age of machine learning-discussion paper},
  author={Van Cranenburgh, Sander and Wang, Shenhao and Vij, Akshay and Pereira, Francisco and Walker, Joan},
  journal={Journal of choice modelling},
  volume={42},
  pages={100340},
  year={2022},
  publisher={Elsevier}
}

@inproceedings{cascetta1996modified,
  title={A modified logit route choice model overcoming path overlapping problems. Specification and some calibration results for interurban networks},
  author={Cascetta, Ennio and Nuzzolo, Agostino and Russo, Francesco and Vitetta, Antonino},
  booktitle={Transportation and Traffic Theory. Proceedings of The 13th International Symposium On Transportation And Traffic Theory, Lyon, France, 24-26 July 1996},
  year={1996}
}

@article{zhonga2023generation,
  title={Generation of aggregated road network by vehicle trajectory data},
  author={Zhong, Hengyi and Seo, Toru and Nakanishi, Wataru and Yasuda, Shohei and Asakura, Yasuo and Iryo, Takamasa},
  journal={EU Science Hub},
  pages={50},
  year={2023}
}

@article{tong2020directed,
  title={Directed graph convolutional network},
  author={Tong, Zekun and Liang, Yuxuan and Sun, Changsheng and Rosenblum, David S and Lim, Andrew},
  journal={arXiv preprint arXiv:2004.13970},
  year={2020}
}

@inproceedings{qiu2025generalized,
  title={Generalized Route Choice Modeling via Fusion of Structural Knowledge and LLM-Inferred Context},
  author={Qiu, Shuhan and Huang, Xiannan and Qin, Guoyang and Chen, Xuejian and Sun, Jian},
  booktitle={2025 IEEE 28th International Conference on Intelligent Transportation Systems (ITSC)},
  pages={1560--1567},
  year={2025},
  organization={IEEE}
}

@article{scarselli2008graph,
  title={The graph neural network model},
  author={Scarselli, Franco and Gori, Marco and Tsoi, Ah Chung and Hagenbuchner, Markus and Monfardini, Gabriele},
  journal={IEEE transactions on neural networks},
  volume={20},
  number={1},
  pages={61--80},
  year={2008},
  publisher={IEEE}
}

@article{bekhor2001stochastic,
  title={Stochastic user equilibrium formulation for generalized nested logit model},
  author={Bekhor, S and Prashker, JN},
  journal={Transportation Research Record},
  volume={1752},
  number={1},
  pages={84--90},
  year={2001},
  publisher={SAGE Publications Sage CA: Los Angeles, CA}
}

@article{he2023stgc,
  title={STGC-GNNs: A GNN-based traffic prediction framework with a spatial--temporal Granger causality graph},
  author={He, Silu and Luo, Qinyao and Du, Ronghua and Zhao, Ling and He, Guangjun and Fu, Han and Li, Haifeng},
  journal={Physica A: Statistical Mechanics and its Applications},
  volume={623},
  pages={128913},
  year={2023},
  publisher={Elsevier}
}

@article{bai2020adaptive,
  title={Adaptive graph convolutional recurrent network for traffic forecasting},
  author={Bai, Lei and Yao, Lina and Li, Can and Wang, Xianzhi and Wang, Can},
  journal={Advances in neural information processing systems},
  volume={33},
  pages={17804--17815},
  year={2020}
}

@article{song2020spatial,
  title     = {Spatial-Temporal Synchronous Graph Convolutional Networks:
               A New Framework for Spatial-Temporal Network Data Forecasting},
  author    = {Song, Chao and Lin, Youfang and Guo, Shengnan and Wan, Huaiyu},
  journal   = {Proceedings of the AAAI Conference on Artificial Intelligence},
  volume    = {34},
  number    = {1},
  pages     = {914--921},
  year      = {2020},
  doi       = {10.1609/aaai.v34i01.5438}
}

@inproceedings{lu2019leveraging,
  title={Leveraging graph neural network with lstm for traffic speed prediction},
  author={Lu, Zhilong and Lv, Weifeng and Xie, Zhipu and Du, Bowen and Huang, Runhe},
  booktitle={2019 IEEE SmartWorld, Ubiquitous Intelligence \& Computing, Advanced \& Trusted Computing, Scalable Computing \& Communications, Cloud \& Big Data Computing, Internet of People and Smart City Innovation (SmartWorld/SCALCOM/UIC/ATC/CBDCom/IOP/SCI)},
  pages={74--81},
  year={2019},
  organization={IEEE}
}

@inproceedings{xie2020deep,
  title={Deep graph convolutional networks for incident-driven traffic speed prediction},
  author={Xie, Qinge and Guo, Tiancheng and Chen, Yang and Xiao, Yu and Wang, Xin and Zhao, Ben Y},
  booktitle={Proceedings of the 29th ACM international conference on information \& knowledge management},
  pages={1665--1674},
  year={2020}
}

@inproceedings{zhong2021probabilistic,
  title={Probabilistic graph neural networks for traffic signal control},
  author={Zhong, Ting and Xu, Zheyang and Zhou, Fan},
  booktitle={ICASSP 2021-2021 IEEE International Conference on Acoustics, Speech and Signal Processing (ICASSP)},
  pages={4085--4089},
  year={2021},
  organization={IEEE}
}

@article{hu2020traffic,
  title={A traffic light dynamic control algorithm with deep reinforcement learning based on GNN prediction},
  author={Hu, Xiaorong and Zhao, Chenguang and Wang, Gang},
  journal={arXiv preprint arXiv:2009.14627},
  year={2020}
}

@article{cheng2026graph,
  title={Graph neural networks for residential location choice: connection to classical logit models},
  author={Cheng, Zhanhong and Hu, Lingqian and Bu, Yuheng and Zhou, Yuqi and Wang, Shenhao},
  journal={Transportation Research Part B: Methodological},
  volume={209},
  pages={103464},
  year={2026},
  publisher={Elsevier}
}

@article{villarraga2025designing,
  title={Designing graph convolutional neural networks for discrete choice with network effects},
  author={Villarraga, Daniel F and Daziano, Ricardo A},
  journal={arXiv preprint arXiv:2503.09786},
  year={2025}
}

@article{hamilton2017inductive,
  title={Inductive representation learning on large graphs},
  author={Hamilton, Will and Ying, Zhitao and Leskovec, Jure},
  journal={Advances in neural information processing systems},
  volume={30},
  year={2017}
}

@article{ZHANG2024103485,
title = {Recursive logit-based meta-inverse reinforcement learning for driver-preferred route planning},
journal = {Transportation Research Part E: Logistics and Transportation Review},
volume = {185},
pages = {103485},
year = {2024},
issn = {1366-5545},
doi = {https://doi.org/10.1016/j.tre.2024.103485},
url = {https://www.sciencedirect.com/science/article/pii/S1366554524000759},
author = {Pujun Zhang and Dazhou Lei and Shan Liu and Hai Jiang}
}

@article{mcfadden2000mixed,
  title={Mixed MNL models for discrete response},
  author={McFadden, Daniel and Train, Kenneth},
  journal={Journal of applied Econometrics},
  volume={15},
  number={5},
  pages={447--470},
  year={2000},
  publisher={Wiley Online Library}
}

@book{mcfadden1977application,
  title={An application of diagnostic tests for the independence from irrelevant alternatives property of the multinomial logit model},
  author={McFadden, Daniel and Tye, William B and Train, Kenneth},
  year={1977},
  publisher={Institute of Transportation Studies, University of California Berkeley}
}

@article{mai2018decomposition,
  title={A decomposition method for estimating recursive logit based route choice models},
  author={Mai, Tien and Bastin, Fabian and Frejinger, Emma},
  journal={EURO Journal on Transportation and Logistics},
  volume={7},
  number={3},
  pages={253--275},
  year={2018},
  publisher={Elsevier}
}

@article{mai2015nested,
  title={A nested recursive logit model for route choice analysis},
  author={Mai, Tien and Fosgerau, Mogens and Frejinger, Emma},
  journal={Transportation Research Part B: Methodological},
  volume={75},
  pages={100--112},
  year={2015},
  publisher={Elsevier}
}

@article{fosgerau2022perturbed,
  title={A perturbed utility route choice model},
  author={Fosgerau, Mogens and Paulsen, Mads and Rasmussen, Thomas Kj{\ae}r},
  journal={Transportation Research Part C: Emerging Technologies},
  volume={136},
  pages={103514},
  year={2022},
  publisher={Elsevier}
}

@article{bovy2008factor,
  title={The factor of revisited path size: Alternative derivation},
  author={Bovy, Piet HL and Bekhor, Shlomo and Prato, Carlo Giacomo},
  journal={Transportation Research Record},
  volume={2076},
  number={1},
  pages={132--140},
  year={2008},
  publisher={SAGE Publications Sage CA: Los Angeles, CA}
}

@incollection{ben1999discrete,
  title={Discrete choice methods and their applications to short term travel decisions},
  author={Ben-Akiva, Moshe and Bierlaire, Michel},
  booktitle={Handbook of transportation science},
  pages={5--33},
  year={1999},
  publisher={Springer}
}

@article{wong2018discriminative,
  title={Discriminative conditional restricted Boltzmann machine for discrete choice and latent variable modelling},
  author={Wong, Melvin and Farooq, Bilal and Bilodeau, Guillaume-Alexandre},
  journal={Journal of choice modelling},
  volume={29},
  pages={152--168},
  year={2018},
  publisher={Elsevier}
}

@article{sifringer2020enhancing,
  title={Enhancing discrete choice models with representation learning},
  author={Sifringer, Brian and Lurkin, Virginie and Alahi, Alexandre},
  journal={Transportation Research Part B: Methodological},
  volume={140},
  pages={236--261},
  year={2020},
  publisher={Elsevier}
}

@article{wong2021reslogit,
  title={ResLogit: A residual neural network logit model for data-driven choice modelling},
  author={Wong, Melvin and Farooq, Bilal},
  journal={Transportation Research Part C: Emerging Technologies},
  volume={126},
  pages={103050},
  year={2021},
  publisher={Elsevier}
}

@inproceedings{he2016identity,
  title={Identity mappings in deep residual networks},
  author={He, Kaiming and Zhang, Xiangyu and Ren, Shaoqing and Sun, Jian},
  booktitle={Computer Vision--ECCV 2016: 14th European Conference, Amsterdam, The Netherlands, October 11--14, 2016, Proceedings, Part IV 14},
  pages={630--645},
  year={2016},
  organization={Springer}
}

@article{tribby2017analyzing,
  title={Analyzing walking route choice through built environments using random forests and discrete choice techniques},
  author={Tribby, Calvin P and Miller, Harvey J and Brown, Barbara B and Werner, Carol M and Smith, Ken R},
  journal={Environment and Planning B: Urban Analytics and City Science},
  volume={44},
  number={6},
  pages={1145--1167},
  year={2017},
  publisher={SAGE Publications Sage UK: London, England}
}

@article{fosgerau2013link,
  title={A link based network route choice model with unrestricted choice set},
  author={Fosgerau, Mogens and Frejinger, Emma and Karlstrom, Anders},
  journal={Transportation Research Part B: Methodological},
  volume={56},
  pages={70--80},
  year={2013},
  publisher={Elsevier}
}

@article{zhao2023deep,
  title={A deep inverse reinforcement learning approach to route choice modeling with context-dependent rewards},
  author={Zhao, Zhan and Liang, Yuebing},
  journal={Transportation Research Part C: Emerging Technologies},
  volume={149},
  pages={104079},
  year={2023},
  publisher={Elsevier}
}

@article{wei2014day,
  title={A Day-to-Day Route Choice Model Based on Reinforcement Learning},
  author={Wei, Fangfang and Ma, Shoufeng and Jia, Ning},
  journal={Mathematical Problems in Engineering},
  volume={2014},
  number={1},
  pages={646548},
  year={2014},
  publisher={Wiley Online Library}
}

@article{dong2022utility,
  title={Utility-based route choice behavior modeling using deep sequential models},
  author={Dong, Guimin and Kweon, Yonghyeon and Park, B Brian and Boukhechba, Mehdi},
  journal={Journal of big data analytics in transportation},
  volume={4},
  number={2},
  pages={119--133},
  year={2022},
  publisher={Springer}
}

@article{wang2021personalized,
  title={Personalized route recommendation with neural network enhanced search algorithm},
  author={Wang, Jingyuan and Wu, Ning and Zhao, Wayne Xin},
  journal={IEEE Transactions on Knowledge and Data Engineering},
  volume={34},
  number={12},
  pages={5910--5924},
  year={2021},
  publisher={IEEE}
}

@article{ben2004route,
  title={Route choice models. Human behaviour and traffic networks},
  author={Ben-Akiva, ME and Scott, MR and Bekhor, Shlomo},
  journal={Springer Berlin Heidelberg},
  pages={23--45},
  year={2004}
}

@article{schmid2022modeling,
  title={Modeling train route decisions during track works},
  author={Schmid, Basil and Becker, Felix and Molloy, Joseph and Axhausen, Kay W and L{\"u}dering, Jochen and Hagen, Julian and Blome, Annette},
  journal={Journal of Rail Transport Planning \& Management},
  volume={22},
  pages={100320},
  year={2022},
  publisher={Elsevier}
}

@article{bekhor2006evaluation,
  title={Evaluation of choice set generation algorithms for route choice models},
  author={Bekhor, Shlomo and Ben-Akiva, Moshe E and Ramming, M Scott},
  journal={Annals of Operations Research},
  volume={144},
  number={1},
  pages={235--247},
  year={2006},
  publisher={Springer}
}

@article{han2020neural,
  title={A neural-embedded choice model: Tastenet-mnl modeling taste heterogeneity with flexibility and interpretability},
  author={Han, Yafei and Pereira, Francisco Camara and Ben-Akiva, Moshe and Zegras, Christopher},
  journal={arXiv preprint arXiv:2002.00922},
  year={2020}
}

@article{yamamoto2002drivers,
  title={Drivers’ route choice behavior: analysis by data mining algorithms},
  author={Yamamoto, Toshiyuki and Kitamura, Ryuichi and Fujii, Junichiro},
  journal={Transportation Research Record},
  volume={1807},
  number={1},
  pages={59--66},
  year={2002},
  publisher={SAGE Publications Sage CA: Los Angeles, CA}
}
\appendix
\section{List of Notation} 
\label{notation}
\begin{table}[htbp]
\centering
\caption{Notations for the RL model and its variants.}
\label{tab:notation_rl}
\small
\setlength{\tabcolsep}{6pt}
\renewcommand{\arraystretch}{1.15}
\begin{tabularx}{\linewidth}{p{0.22\linewidth} X}
\hline
\textbf{Symbol} & \textbf{Definition} \\
\hline
$G=\langle\mathcal{A},\mathcal{V}\rangle$ & Road network graph; $\mathcal{A}$ is the set of nodes and $\mathcal{V}$ is the set of links. \\
$A$ & Adjacency matrix of links. \\
$A_{ka}$ & Entry of $A$; $A_{ka}=1$ if link $a$ is reachable from link $k$. \\
$\delta(a\mid k)$ & Feasibility indicator; equals $1$ if $A_{ka}=1$, otherwise $0$. \\

$O(k)$ & Set of outgoing links from link $k$. \\
$u^{\mathrm{RL}}(a\mid k;\beta)$ & Instantaneous utility of choosing $a$ from $k$ in RL. \\
$v^{\mathrm{RL}}(a\mid k;\beta)$ & Deterministic component of instantaneous utility in RL. \\
$\phi$ & Parameters of the systematic utility specification. \\
$\epsilon(a)$ & i.i.d. extreme value type-I error term for link $a$. \\
$\mu$ & Scale parameter of the Gumbel error. \\
$V^d(k)$ & Expected maximum downstream utility when the destination is $d$. \\
$P^d(a\mid k)$ & Choice probability of choosing link $a$ from $k$ with destination $d$. \\
$P^d$ & Choice probability matrix with entries $P^d_{ka}=P^d(a\mid k)$. \\
$I$ & Identity matrix. \\
$G^a$ & A vector whose $a$-th element is $1$ and others are $0$. \\
$F^{od}$ & OD-specific expected link flow (used as link-size attribute in LS-RL). \\
LS-RL, NRL & Link-size Recursive Logit; Nested Recursive Logit. \\
\hline
\end{tabularx}
\end{table}
\begin{table}[htbp]
\centering
\caption{Notations for Res-RL.}
\label{tab:notation_resrl}
\small
\setlength{\tabcolsep}{6pt}
\renewcommand{\arraystretch}{1.15}
\begin{tabularx}{\linewidth}{p{0.22\linewidth} X}
\hline
\textbf{Symbol} & \textbf{Definition} \\
\hline

$u^{r}(a\mid k)$ & Instantaneous utility of choosing link $a$ from link $k$ in Res-RL. \\
$v^{r}(a\mid k)$ & Systematic component of $u^{r}(a\mid k)$ in Res-RL; specified as a linear function of action features. \\
$g^{r}(a\mid k)$ & Residual component of $u^{r}(a\mid k)$ in Res-RL. \\
$M$ & Number of residual layers. \\
$h^{r,m}$ & $m$-th hidden layer of Res-RL. \\
$h^{r,0}_{ka}$ & The input layer of Res-RL, which is defined as $h^{r,0}_{ka}=v^{r}(a\mid k)$. \\
$\theta^{r,m}$ & Weight matrix of the $m$-th residual layer. \\
$G^{r}$ & Residual component matrix whose $(k,a)$ entry corresponds to $g^{r}(a\mid k)$. \\
$\mathbf{1}_{|\mathcal{V}|\times|\mathcal{V}|}$ & All-ones matrix used in the residual-layer transformation. \\
$\ln(\cdot)$, $\exp(\cdot)$ & Element-wise natural logarithm and exponential operators in the residual layers. \\
$\odot$ & Hadamard (element-wise) product used to mask infeasible transitions. \\

\hline
\end{tabularx}
\end{table}
\begin{table}[htbp]
\centering
\caption{Notations for the ResDGCN-RL model.}
\label{tab:notation_resdgcnrl}
\small
\setlength{\tabcolsep}{6pt}
\renewcommand{\arraystretch}{1.15}
\begin{tabularx}{\linewidth}{p{0.22\linewidth} X}
\hline
\textbf{Symbol} & \textbf{Definition} \\
\hline
$u^{d}(a\mid k)$ & Instantaneous utility of choosing link $a$ from link $k$ in ResDGCN-RL (destination-specific). \\
$v^{d}(a\mid k)$ & Systematic component of $u^{d}(a\mid k)$ in ResDGCN-RL. \\
$g^{d}(a\mid k)$ & Residual component of $u^{d}(a\mid k)$ in ResDGCN-RL learned by the modified DGCN. \\
$G^{d}$ & Residual component matrix with entries $G^{d}(k,a)=g^{d}(a\mid k)$. \\
$h^{d,m}$ & $m$-th hidden layer of ResDGCN-RL. \\
$h^{d,0}_{k,a}$ & The input layer of ResDGCN-RL, which is defined as $h^{d,0}_{k,a}=v^{d}(a\mid k)$. . \\
$\theta^{d,m}$ & Weight matrix in the $m$-th hidden layer of ResDGCN-RL. \\
$F(\cdot)$ & Modified DGCN propagation rule used in the residual component. \\
$Z_F$ & First-order proximity matrix. \\
$Z_{S_{\textrm{in}}}$ & Second-order in-degree proximity matrix. \\
$Z_{S_{\textrm{out}}}$ & Second-order out-degree proximity matrix. \\
$\alpha,\beta,\gamma$ & Learnable scalars weighting the contributions of $Z_F$, $Z_{S_{\textrm{in}}}$, and $Z_{S_{\textrm{out}}}$, respectively. \\
\hline
\end{tabularx}
\end{table}


\section{Importance of parameters $\alpha$, $\beta$ and $\gamma$}
\label{subs2}

This appendix provides an illustrative example to clarify the role of the first and second-order proximity parameters in ResDGCN-RL, namely $\alpha$, $\beta$, and $\gamma$. These parameters act as scaling coefficients that control how different proximity structures contribute to the cross-effect encoded in the residual utility. Thw network used for illustration is showed in Figure \ref{networkc4}.

\begin{figure}[htbp]
  \centering
  \includegraphics[width=400pt]{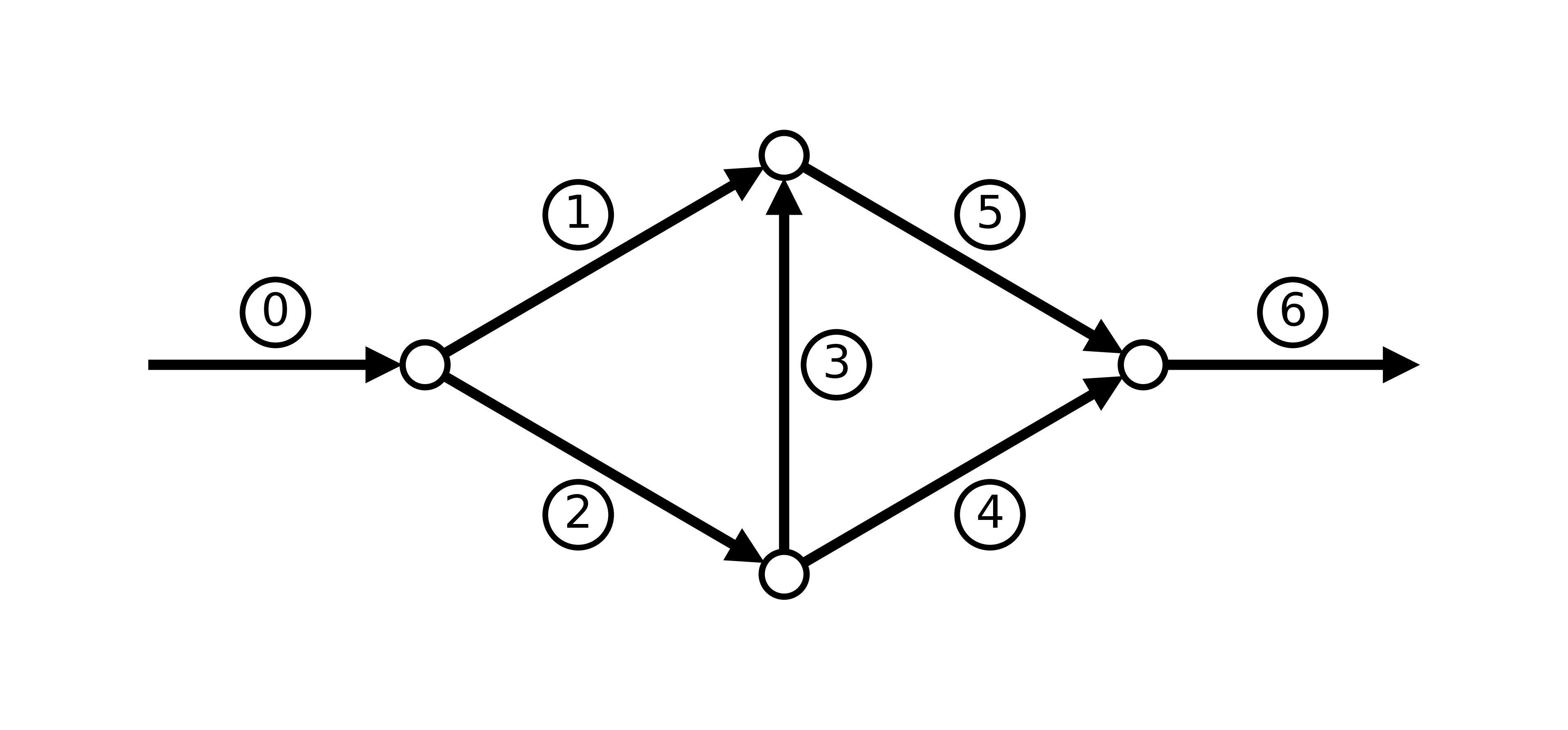}
  \caption{A very simple 3-path network for Appendix B.}
  \label{networkc4}
\end{figure}

In this example, a one-layer ResDGCN-RL is considered for simplicity and focus on the parameter $\theta^{d,0}_{34}$, which in fact encompasses two distinct meanings. Specifically, the partial derivatives of $g^d(4 \mid 3)$ and $g^d(4 \mid 1)$ with respect to $v^d(3 \mid 2)$ are given as follows:

\begin{equation}
    \label{g53}
    \frac{\partial g^d(4\mid3)}{\partial v^d(3\mid2)}=\frac{1-2e^{-g^d(4\mid3)}}{2e^{-g^d(4\mid3)}}\alpha Z_f(3,2)\theta^{d,0}_{34}
\end{equation}

\begin{equation}
    \label{g51}
    \frac{\partial g^d(4\mid1)}{\partial v^d(3\mid2)}=\frac{1-2e^{-g^d(4\mid1)}}{2e^{-g^d(4\mid1)}}\gamma Z_{\textrm{out}}(1,2)\theta^{d,0}_{34}
\end{equation}

In this case, $\theta^{d,0}_{34}$ reflects the cross-effect of traveling from link 2 to link 3 on two actions: from link 3 to link 4 and from link 1 to link 4. These effects arise from different types of proximity, namely first-order proximity and second-order out-degree proximity. More generally, a single element in $\theta$ can incorporate cross-effect stemming from multiple proximity types. Within this framework, $\alpha$, $\beta$, and $\gamma$ function as adjustment coefficients that account for the contributions of each proximity type. 


\section{Gradient formulation}
\label{gradient}
The parameter vectors $\phi$ (systematic component) and $\theta$ (residual layers) is estimated by maximizing the log-likelihood of observed link choices. For Res-RL with $M$ hidden layers, the analytical derivative with respect to $\beta_q$ is defined as:
\begin{equation}
    \frac{\partial \textrm{Loss}}{\partial\phi_q}
    = -\frac{1}{\mu}\sum_{n=1}^N\left[\sum_{l=1}^{l_n-1}\frac{\partial h^{r,M}_{a_{l}^n a_{l+1}^n}}{\partial \phi_q}
    - \frac{\partial V^{a^n_{l_n}}(a_0^n)}{\partial \phi_q}\right]
\end{equation}

Based on the conclusion of \citet{fosgerau2013link}, the term $\frac{\partial V_{a_0^n}}{\partial \phi_q}$ depends only on $\frac{\partial h^{r,M}}{\partial \phi_q}$:
\begin{equation}
    \frac{\partial V^{a^n_{l_n}}(a_0^n)}{\partial \phi_q}
    = \mu e^{-\frac{1}{\mu}V^{a^n_{l_n}}(a_0^n)}
    \left[(I-M)^{-1}e^{h^{r,M}}
    \left[\frac{\partial h^{r,M}}{\partial \phi_q}\right]
    e^{\frac{1}{\mu}V^{a^n_{l_n}}}\right]_{a_0^n}
\end{equation}
where $M=\exp(h^{r,M})\otimes A$.

This implies that, in practice, it is sufficient to compute the gradient of the deterministic component of the instantaneous utility, $h^{r,M}$, with respect to the model parameters, without explicitly differentiating through the matrix inversion.

Let
\begin{equation}
    f^{r,m}(h^{r,m-1}) = -\ln\left(\mathbf{1}_{\lvert \mathcal{V} \rvert \times \lvert \mathcal{V} \rvert}+\exp\left(h^{r,m-1}\theta^{r,m}\right)\right).
\end{equation}

The gradient of $h^{r,M}_{ka}$ with respect to the parameter for $q$-th feature $\beta_q$ is given by:
\begin{equation}
     \frac{\partial h^{r,m}_{ka}}{\partial \phi_q}
     = \frac{\partial h^{r,0}_{ka}}{\partial \phi_q}
     + \sum_{m'=1}^{m}\frac{\partial f^{r,m'}_{ka}}{\partial \phi_q},
\end{equation}
where
\begin{equation}
    \frac{\partial h^{r,0}_{ka}}{\partial \phi_q} = x_q(a|k),
\end{equation}
where $x_q(a|k)$ is the $q$-th feature of traveling from $k$ to a, and
\begin{equation}
\frac{\partial f^{r,m}_{ka}}{\partial \phi_q}
=
\frac{\exp\!\Big(\sum_{a'} h^{r,m-1}_{k a'}\,\theta^{r,m}_{a'a}\Big)}
{1+\exp\!\Big(\sum_{a'} h^{r,m-1}_{k a'}\,\theta^{r,m}_{a'a}\Big)}
\sum_{a''}\theta^{r,m}_{a''a}\,\frac{\partial h^{r,m-1}_{k a''}}{\partial \phi_q}
\end{equation}

The same recursive principle extends to the Res-RL weight matrix as well as to the parameters in ResDGCN-RL.

Building on the preceding derivations, the proposed model exhibits a skip-connection mechanism \citep{he2016identity} that mitigate vanishing/exploding gradients in Res-RL and ResDGCN-RL and preventing degradation as depth increases; 
\begin{equation}
\label{skip}
\begin{aligned}
  \frac{\partial h^{r,M}}{\partial \beta}=\frac{\partial u^r(a|k)}{\partial v^r(a|k)}\times\frac{\partial v^r(a|k)}{\partial \phi}+&\frac{\partial h^{r,M}}{\partial h^{r,1}}\times \frac{\partial h^{r,1}}{\partial \phi}+\ldots+\\
  &\frac{\partial h^{r,M}}{\partial h^{r,M-1}}\times \frac{\partial h^{r,M-1}}{\partial \phi}
  \end{aligned}
\end{equation}

Equation \eqref{skip} shows the nature of skip-connection, where the derivative of residual layers is independently computed. Each term in the summation corresponds to an independent gradient path, meaning that even if one term approaches zero, the total gradient remains nonzero, allowing effective learning.

In contrast, if the residual connection is removed—meaning that each layer’s output only depends on the computations from the previous layer—the gradient must be propagated using the chain rule:
\begin{equation}
\label{chain}
  \frac{\partial h^{r,M}}{\partial \phi}=\frac{\partial h^{r,M}}{\partial h^{r,M-1}}\times\frac{\partial h^{r,M-1}}{\partial h^{r,M-2}}\times\frac{\partial h^{r,M-2}}{\partial h^{r,M-3}}\times \ldots \times
  \frac{\partial h^{r,1}}{\partial v^r(a|k)}\times\frac{\partial v^r(a|k)}{\partial \phi}
\end{equation}

As Equation \eqref{chain} shows, if any intermediate derivative is zero, the overall gradient collapses to zero, so $\beta$ cannot be learned and the model becomes non-identifiable.

{
\section{Road network aggregation}
\label{road network aggregation}
The aggregated road network is generated from vehicle trajectory data. The aggregation method consists of three main steps: node detection, link generation, and network reorganization.

First, the trajectory data are used to detect nodes. Since nodes in a road network generally represent intersections, vehicles may exhibit turning movements in these areas. Accordingly, intersection areas are identified as spatial clusters of trajectory points associated with large turning angles. Each identified intersection area is then represented by a single aggregated node.

Second, each trajectory point is associated with a pair of nodes: the last node passed by the vehicle and the next node subsequently reached. A directed link is generated between two nodes when trajectory observations indicate that vehicles travel from one node to the other.

Third, the road network is reorganized according to traffic volume. Links with traffic volumes below a specified threshold are removed, allowing the level of network aggregation to be controlled. For the road network used in this study, the threshold is set to 100 probe vehicles per day.

\section{Estimation result of temporal and destination holdout}
\label{holdout}
\begin{table}[htbp]
\centering
\setlength{\tabcolsep}{3pt} 
\fontsize{8pt}{10pt}\selectfont
\caption{Parameter estimates (standard errors in parentheses) and model performance comparison across models of temporal holdout . }

\label{tab:param_estimates_th}
\begin{tabular}{lccccc}
\toprule
Parameter 
 & RL 
 & RL-LS 
 & NRL 
 & \makecell{Res-RL\\$(M=2,\lambda=0)$} 
 & \makecell{ResDGCN-\\RL $(M=2,\lambda=0)$} 
 \\
\midrule
$\beta_{\mathrm{TT}}$  
 & \makecell{-0.745 \\ (0.003)} 
 & \makecell{ -0.788 \\ (0.002)} 
 & \makecell{-0.762 \\ (0.005)} 
 & \makecell{-0.603 \\ (0.006)} 
 & \makecell{-0.635 \\ (0.005)} 
 \\
$\beta_{\mathrm{Spd}}$ 
 & \makecell{0.124 \\ (0.010)} 
 & \makecell{0.221 \\ (0.005)} 
 & \makecell{0.396 \\ (0.007)} 
 & \makecell{-0.007$^{*}$  \\ ( 0.023)} 
 & \makecell{0.026 \\ (0.008)} 
 \\
$\beta_{\mathrm{RT}}$  
 & \makecell{-1.647 \\ (0.004)} 
 & \makecell{-1.733 \\ (0.004)} 
 & \makecell{-1.655 \\ (0.010)} 
 & \makecell{-1.605 \\ (0.008)} 
 & \makecell{-1.606 \\ (0.008)} 
 \\
$\beta_{\mathrm{LC}}$  
 & \makecell{-0.181 \\ (0.007)} 
 & \makecell{-0.152 \\ (0.004)} 
 & \makecell{-0.318 \\ (0.006)} 
 & \makecell{-0.136 \\ (0.016)} 
 & \makecell{-0.136 \\ (0.004)} 
\\
$\beta_{\mathrm{LS}}$  
 & -- 
 & \makecell{-0.170 \\ (0.004)} 
 & -- & -- & --  \\
$\gamma_{\mathrm{TT}}$ 
 & -- & -- 
 & \makecell{0.121 \\ (0.005)} 
 & -- & -- \\
$\gamma_{\mathrm{Spd}}$ 
 & -- & -- 
 & \makecell{-0.203 \\ (0.002)} 
 & -- & --  \\
$\gamma_{\mathrm{LS}}$  
 & -- & -- 
 & \makecell{-0.020 \\ (0.003)} 
 & -- & --  \\
\midrule
$\alpha$ 
 & -- & -- & -- & -- & 1.046  \\
$\beta$ 
 & -- & -- & -- & -- & 1.056 \\
$\gamma$ 
 & -- & -- & -- & -- & 1.080 \\
\midrule
Log-likelihood         
 & -228799.152 & -207503.556 & -191011.388 & -86670.858 & -84407.289 \\
ACP                    
 & 0.526      & 0.534      & 0.537      & 0.762      & 0.769       \\
JSD                    
 & 0.032      & 0.028      & 0.026      & 0.013      & 0.012       \\
BLEU                   
 & 0.732      & 0.738      & 0.743      & 0.846      & 0.849       \\
Number of parameters   
 & 4          & 5          & 7          & 7158       & 77569      \\
\bottomrule
\end{tabular}

\vspace{2pt}
\begin{minipage}{0.95\linewidth}
\footnotesize $^{*}$\,p$\geq$0.01.
\end{minipage}
\end{table}

\begin{table}[htbp]
\centering
\setlength{\tabcolsep}{3pt} 
\fontsize{8pt}{10pt}\selectfont
\caption{Parameter estimates of destination holdout(standard errors in parentheses) and model performance comparison across models. }

\label{tab:param_estimates_dh}
\begin{tabular}{lccccc}
\toprule
Parameter 
 & RL 
 & RL-LS 
 & NRL 
 & \makecell{Res-RL\\$(M=2,\lambda=0)$} 
 & \makecell{ResDGCN-\\RL $(M=2,\lambda=0)$} 
 \\
\midrule
$\beta_{\mathrm{TT}}$  
 & \makecell{-0.740 \\ (0.002)} 
 & \makecell{ -0.801 \\ (0.004)} 
 & \makecell{-0.777 \\ (0.005)} 
 & \makecell{-0.582 \\ (0.002)} 
 & \makecell{-0.578 \\ (0.003)} 
 \\
$\beta_{\mathrm{Spd}}$ 
 & \makecell{0.162 \\ (0.008)} 
 & \makecell{0.161 \\ (0.006)} 
 & \makecell{0.120 \\ (0.004)} 
 & \makecell{-0.014 \\ ( 0.003)} 
 & \makecell{0.010$^{*}$ \\ (0.011)} 
 \\
$\beta_{\mathrm{RT}}$  
 & \makecell{-1.545 \\ (0.003)} 
 & \makecell{-1.733 \\ (0.004)} 
 & \makecell{-1.347 \\ (0.011)} 
 & \makecell{-1.617 \\ (0.006)} 
 & \makecell{-1.585 \\ (0.006)} 
 \\
$\beta_{\mathrm{LC}}$  
 & \makecell{-0.171 \\ (0.006)} 
 & \makecell{-0.152 \\ (0.004)} 
 & \makecell{-0.328 \\ (0.009)} 
 & \makecell{-0.141 \\ (0.004)} 
 & \makecell{-0.106 \\ (0.009)} 
\\
$\beta_{\mathrm{LS}}$  
 & -- 
 & \makecell{-0.152 \\ (0.003)} 
 & -- & -- & --  \\
$\gamma_{\mathrm{TT}}$ 
 & -- & -- 
 & \makecell{0.128 \\ (0.014)} 
 & -- & -- \\
$\gamma_{\mathrm{Spd}}$ 
 & -- & -- 
 & \makecell{-0.045 \\ (0.010)} 
 & -- & --  \\
$\gamma_{\mathrm{LS}}$  
 & -- & -- 
 & \makecell{-0.021 \\ (0.004)} 
 & -- & --  \\
\midrule
$\alpha$ 
 & -- & -- & -- & -- & 1.082  \\
$\beta$ 
 & -- & -- & -- & -- & 1.059 \\
$\gamma$ 
 & -- & -- & -- & -- & 1.100 \\
\midrule
Log-likelihood         
 & -86651.288 & -81300.672 & -79102.323 & -35195.333 & -34887.408 \\
ACP                    
 & 0.485      & 0.501      & 0.511      & 0.673      & 0.678       \\
JSD                    
 & 0.036      & 0.032      & 0.031      & 0.018      & 0.017       \\
BLEU                   
 & 0.750      & 0.751      & 0.754      & 0.836      & 0.837       \\
Number of parameters   
 & 4          & 5          & 7          & 7158       & 77569      \\
\bottomrule
\end{tabular}

\vspace{2pt}
\begin{minipage}{0.95\linewidth}
\footnotesize $^{*}$\,p$\geq$0.01.
\end{minipage}
\end{table}

}

\end{document}